\documentclass[journal,10pt]{IEEEtran}
\IEEEoverridecommandlockouts

\usepackage{cite}
\usepackage[colorlinks=true, linkcolor=blue, citecolor=blue, urlcolor=blue,]{hyperref}

\usepackage{booktabs}
\usepackage{threeparttable}

\usepackage{pifont}
\newcommand{\cmark}{\textcolor{green}{\ding{51}}} % ✓
\newcommand{\xmark}{\textcolor{red}{\ding{55}}}   % ✗

\usepackage{amsmath,amssymb,amsfonts}
\usepackage{algorithm}
\usepackage{algpseudocode}

\usepackage{graphicx}
\usepackage{multirow}
\usepackage{textcomp}
\usepackage{xcolor}
\def\BibTeX{{\rm B\kern-.05em{\sc i\kern-.025em b}\kern-.08em
    T\kern-.1667em\lower.7ex\hbox{E}\kern-.125emX}}

\begin{document}

\title{Winning the Lottery by Preserving Network Training Dynamics with Concrete Ticket Search}
 
\author{Tanay~Arora and Christof~Teuscher,~\IEEEmembership{Senior Member, ~IEEE}
}

\maketitle

\begin{abstract}
The Lottery Ticket Hypothesis asserts the existence of highly sparse, trainable subnetworks (\textit{`winning tickets'}) within dense, randomly initialized neural networks. However, state-of-the-art methods of drawing these tickets, like Lottery Ticket Rewinding (LTR), are computationally prohibitive, while more efficient saliency-based Pruning-at-Initialization (PaI) techniques suffer from a significant accuracy-sparsity trade-off and fail basic sanity checks. In this work, we argue that PaI's reliance on first-order saliency metrics, which ignore inter-weight dependencies, contributes substantially to this performance gap, especially in the sparse regime. To address this, we introduce Concrete Ticket Search (CTS), an algorithm that frames subnetwork discovery as a holistic combinatorial optimization problem. By leveraging a Concrete relaxation of the discrete search space and a novel gradient balancing scheme (\textsc{GradBalance}) to control sparsity, CTS efficiently identifies high-performing subnetworks near initialization without requiring sensitive hyperparameter tuning. Motivated by recent works on lottery ticket training dynamics, we further propose a knowledge~distillation-inspired family of pruning objectives, finding that minimizing the reverse Kullback-Leibler divergence between sparse and dense network outputs (CTS\textsubscript{KL}) is particularly effective. Experiments on varying image classification tasks show that CTS produces subnetworks that robustly pass sanity checks and achieve accuracy comparable to or exceeding LTR, while requiring only a small fraction of the computation. For example, on ResNet-20 on CIFAR10, CTS\textsubscript{KL} produces subnetworks of \(99.3\%\) sparsity with a top-1 accuracy of \(74.0\%\) in just \(7.9\) minutes, while LTR produces subnetworks of the same sparsity with an accuracy of \(68.3\%\) in \(95.2\) minutes. However, while CTS outperforms saliency-based methods in the sparsity-accuracy tradeoff across all sparsities, such advantages over LTR emerge most clearly only in the highly sparse regime.

\end{abstract}


\section{Introduction}
\IEEEPARstart{O}{ver} the last decades, the exponential scaling of neural networks and the demand for edge deployment have led significant research into model compression through \textit{pruning}, i.e., removing a large portion of their weights. Although traditionally implemented after training due to conventional knowledge that overparameterization was necessary for effective gradient descent \cite{overparameterizationgeneralization}, \cite{overparameterizationacceleration}, Frankle and Carbin \cite{lotterytickethypothesis} provide empirical evidence that this computationally extensive process can be avoided. 

They propose the \textit{Lottery Ticket Hypothesis (LTH):} over any task, given a sufficiently complex, randomly initialized model, there exist \textit{sparse subnetworks that can be trained to accuracy comparable to that of their fully trained dense counterparts}. Termed ``winning'' or ``lottery'' tickets, they find these subnetworks through Iterative Magnitude Pruning (IMP), which involves a cycle of training to convergence, pruning a small percentage of least-magnitude weights, and `rewinding' the remaining weights to their value at initialization. In practice, however, on deeper models, IMP is unable to find lottery tickets unless rewound to an iteration slightly after initialization, which is known as Lottery Ticket Rewinding (LTR) \cite{instability}.

The intense retraining required by LTR defeats the original purpose of lottery tickets: to improve training efficiency. To address this, researchers have explored pruning methods that identify such winning tickets at the start of training, known as Pruning-at-Initialization (PaI) methods. However, PaI methods consistently fall short of LTR, showing a noticeable tradeoff between sparsity and accuracy. Frankle and Carbin \cite{pruningatinitialization} demonstrated that popular PaI methods like SNIP, GraSP, and SynFlow also fail basic sanity checks, such as randomly shuffling the pruning mask within each layer. Such methods seem to merely identify good sparsity ratios for each layer, but lose initialization-specific information. They remark that the consistent failure of these diverse methods may point to a fundamental weakness in pruning at initialization and urge the development of new signals to guide pruning early in training. 

To our knowledge, the search for an efficient algorithm to draw lottery tickets near initialization with state-of-the-art accuracy remains an open problem.

In this work, we answer this challenge by introducing Concrete Ticket Search (CTS), which reframes ticket discovery as a direct combinatorial optimization problem. Rather than scoring weight importance independently, CTS learns entire tickets through a concrete relaxation, \cite{concretedistribution}, of the binary mask space. A novel gradient balancing scheme ensures that the target sparsity is met without any extensive tuning, and pruning objectives inspired by knowledge distillation rather than naive task loss guide the sparse model to preserve the behavior of its dense counterpart. For example, an especially effective objective comes from minimizing the reverse Kullback-Leibler (KL) divergence of the ticket with respect to its dense parent.

The main contributions of this work can be summarized as follows. We shed light on a fundamental flaw in the first-order approximations used by nearly all PaI works. As such, we posit that holistic subnetwork search is a necessity and develop CTS, an algorithm that draws lottery tickets near initialization with accuracy better than or comparable to the current state-of-the-art, LTR. It avoids the need for extra hyperparameter tuning and achieves these results in significantly less time. For example, at \(\mathbin{\sim} 200\) times compression (\(0.47\%\) density), the proposed method draws tickets in 192 (Quick CTS) and 24 (CTS) times less training iterations than LTR on CIFAR-10 tasks; this speedup only increases as subnetworks grow more and more sparse. These lottery tickets pass sanity checks as proposed in \cite{pruningatinitialization} and \cite{sanitychecks}. Further, inspired by knowledge distillation, we propose pruning criteria that perform better than existing approaches when pruning early in training.

\begin{table*}
\centering
\begin{threeparttable}
\caption{Comparison of different pruning methods for ResNet-20 on CIFAR-10 at 1.44\% density, with rows corresponding to CTS bolded. We obtain test-accuracy results for Edge-Popup and Gem-Miner from \cite{gemminer}}
\label{tab:methodcomparison}
\begin{tabular}{lcccc}
\toprule
Pruning Method & Passes Sanity Checks & Avoids Hyperparameter Tuning & Computation Required & Test Accuracy (\%) \\
\midrule
LTR\tnote{a} ~\cite{instability} & \cmark & \cmark & $3058$ epochs & $80.90$ \\
SNIP~\cite{snip}  & \xmark & \cmark & $160$ epochs & $67.73$ \\
GraSP~\cite{grasp}  & \xmark & \cmark & $160$ epochs & $62.59$ \\
SynFlow~\cite{synflow} & \xmark & \cmark & $161$ epochs & $70.18$\\
Edge-popup~\cite{edgepopup}  & \xmark & \cmark & $320$ epochs & $10.00$\\
Gem-Miner\tnote{b} ~\cite{gemminer}  & \cmark & \xmark & $320$ epochs & $77.89$ \\
\textbf{Quick \(\textsc{CTS}_{\text{KL}}\)}\tnote{a} & \cmark & \cmark & $180$ epochs & $\boldsymbol{79.04}$ \\
\textbf{\(\textsc{CTS}_{\text{KL}}\)}\tnote{a} & \cmark & \cmark & $320$ epochs & $\boldsymbol{80.26}$ \\
\bottomrule
\end{tabular}
\begin{tablenotes}
\footnotesize
\item[a] Rewinding iteration used is k=3000, i.e., epoch 7.5. For more details, see discussion in Section~\ref{ss:experimentalconfig}.
\item[b] The authors tune training hyperparameters extensively. This may be a cause of slight discrepancies.
\end{tablenotes}
\end{threeparttable}
\end{table*}

\section{Related Work}

While there have been a variety of approaches explored to compress neural networks, in this work, we focus on unstructured pruning.
\subsection{Pruning During / After Training}

Traditionally, network pruning has been treated as a post-processing step applied to fully trained, dense models. The prevailing wisdom was that the overparameterization of dense networks was essential for successful optimization dynamics \cite{overparameterizationgeneralization}, \cite{overparameterizationacceleration}. The most common methods operate by assigning an importance score, or saliency score, to each weight and removing those with the lowest scores. Common criteria include least-magnitude pruning \cite{han2015compression}, and those that leverage second-order information of the loss landscape, as in \cite{optimalbrainsurgeon}, \cite{chitaprune}, and \cite{snowsprune}. However, one-shot approaches suffer from extreme drops in performance in the sparse regime, often due to layer collapse: all weights from a single layer are removed and the network is rendered naive. To combat this, effective works all follow a taxing iterative procedure consisting of repeatedly pruning a small percentage of weights and reconstructing weights through finetuning \cite{learningraterewinding}, \cite{whylearningraterewinding}, \cite{spodeprune}.

An alternative paradigm features gradual pruning during training, incorporating sparsity-inducing regularization into the training objective. For example, \(L_1\) regularization \cite{proxsgd} encourages weights to become exactly zero, while other works explicitly regularize the \(L_0\) norm using continuous relaxations \cite{learningl0sparse}.

Although these methods are able to compress neural networks for inference, there is no reduction in the cost of training.

\subsection{Pruning At Initialization}

The landscape of network pruning was reshaped by Frankle and Carbin \cite{lotterytickethypothesis} with the LTH, where they demonstrated the existence of lottery tickets. These tickets, when trained in isolation, can match or exceed the performance of the original dense network trained for the same number of iterations. However, the SOTA method for drawing lottery tickets, LTR, requires an extremely extensive retraining process. To our knowledge, LTR \textit{with sufficiently late rewinding} remains SOTA in terms of the sparsity-accuracy tradeoff over all sparsities for all pruning methods, including those that occur during or after training.

Motivated by the inefficiency of IMP, a significant body of research has focused on PaI methods that prune before training. These methods typically maintain the paradigm of calculating saliency scores for each weight. For example, SNIP measures the effect of each weight on the loss function \cite{snip}, GraSP aims to preserve the gradient flow throughout the network by examining the Hessian-gradient product \cite{grasp}, and SynFlow preserves synaptic flow \cite{synflow}.

Despite their efficiency, PaI methods consistently drastically underperform LTR. Their systematic failure on sanity checks (e.g., reinitialization, layerwise shuffling, etc.) further suggests that their initialization-based, first-order saliency scores do not capture the necessary conditions for successful training \cite{pruningatinitialization}, \cite{sanitychecks}. While newer works, e.g., \cite{patheXclusion}, \cite{epsd}, somewhat improve upon the performance of these PaI methods, none are able to rival LTR in terms of sparsity-accuracy performance. Like our work, the first of these, path eXclusion~\cite{patheXclusion}, aims to address the interdependence of weights through the neural tangent kernel. However, this approximation of the model as linear is unable to sufficiently avoid the complex interdependency of the weights due to factors like batch normalization.

\subsection{Supermasks}

A related line of work explores learning binary ``supermasks'' for fixed, randomly initialized networks \cite{supermasks}. In this paradigm, subnetworks of good performance are found at initialization, and then possibly further finetuned. Edge-popup \cite{edgepopup}, for example, assigns each weight a score and uses a straight-through estimator to learn a mask with fixed layerwise-sparsities, though its outputs are not susceptible to further finetuning. More recent methods like Gem-Miner \cite{gemminer} improve upon edge-popup through avoiding a strict layerwise sparsity pattern and adding sparsity regularization techniques. These works are similar to ours in the sense that they also implement holistic subnetwork optimization over a fixed network. However, they are only designed to learn good task performance immediately rather than search for genuine lottery tickets, and the use of a straight-through estimator can bias and slow down optimization. Additionally, these methods often introduce a new set of sensitive hyperparameters for the mask-learning process, which can be difficult to tune. 

\section{Limitations of the First-Order Saliency Approach} \label{s:issuewithsaliency}

Nearly all PaI methods, as well as numerous non-PaI approaches, rely on a class of gradient-based metrics to assign importance scores to individual weights, enabling the pruning of the least important ones. Let \(\mathcal{R}\) denote a scalar objective function -- not necessarily task loss -- that we aim to minimize. Then, this family of such metrics, known as synaptic saliency, can be expressed as the element-wise product:
\begin{equation}
\mathcal{S}(\theta) = \frac{\partial \mathcal{R}}{\partial \theta} \odot \theta,
\label{saliencydefinition}    
\end{equation}
where \(\theta\) parametrizes the network. This saliency-based pruning is equivalent to removes weights whose elimination most reduces the first-order approximation of \(\mathcal{R}\). For pruning weight \(\theta_j\), define the perturbation \(\delta = -\theta \odot e_j\), where \(e_j\) is the \(j\)-th standard basis vector. Then,
\[
\Delta \mathcal{R} \approx (\nabla_{\theta} \mathcal{R})^{\top} \, \delta \theta = -\frac{\partial R}{\partial \theta_j} \theta_j.
\]
This first-order approach only considers the effect of removing weights independently and ignores interactions between weights. However, it has been observed that network weights are highly coupled, particularly during training. 

For example, under the study of neural circuits -- computational subgraphs within a network responsible for a specific behavior -- it has been shown that many circuits are responsible for translated versions of the same basic feature \cite{equivariantcircuits}. Brokman et al., \cite{correlationmodedecomp}, show that network parameters can be decomposed into correlation modes, which have synchronized updates in state space. As architectures become more and more deep and complex non-nonlinearities like batch normalization remain present, such interactions between weights only becomes more difficult to model or predict.

\begin{figure}
    \centering
    \includegraphics[width=\linewidth]{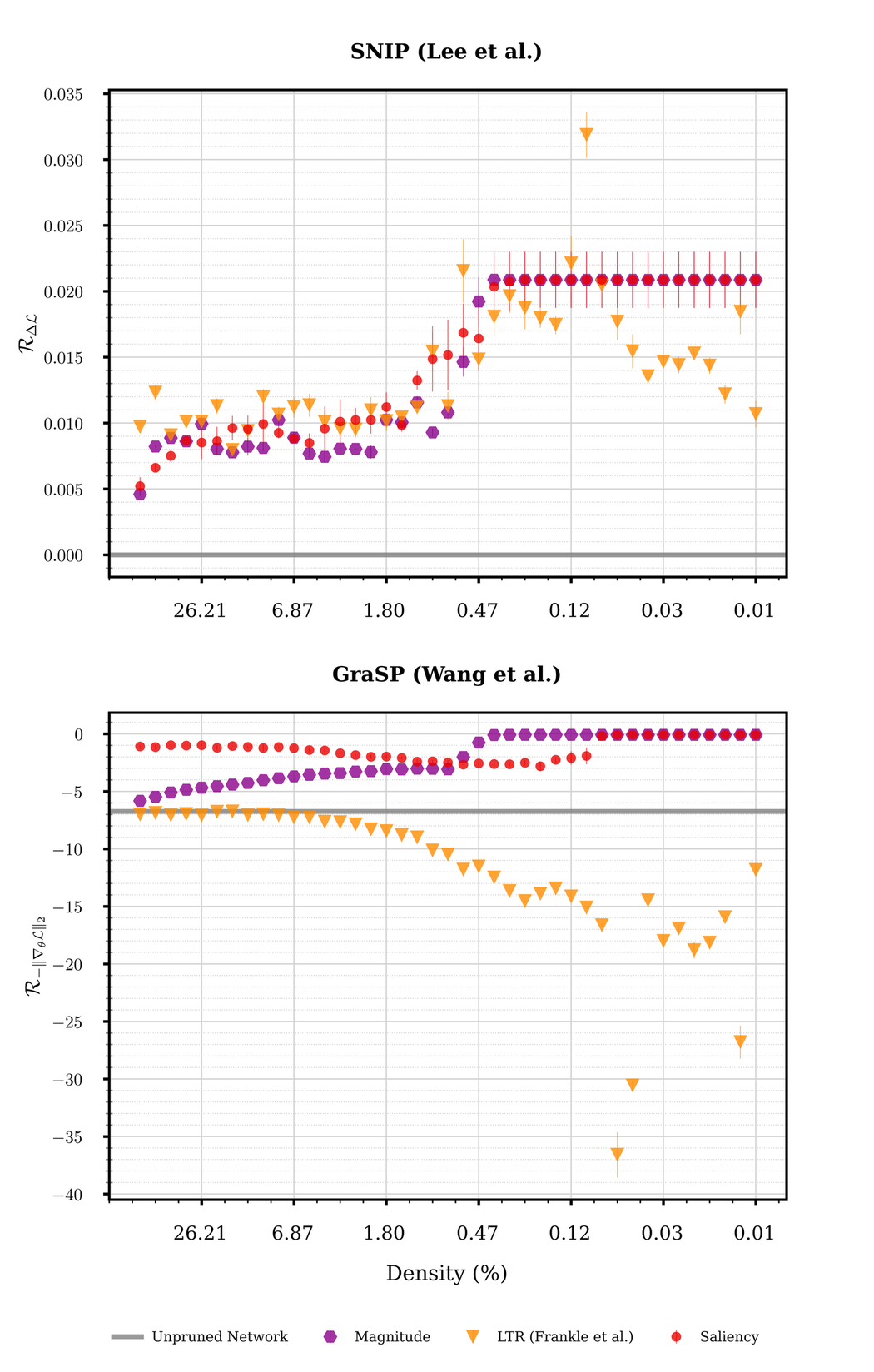}
    \caption{Performance on corresponding objective functions of saliency-based pruning in comparison to LTR and least-magnitude pruning. As described in sections \ref{s:issuewithsaliency} and \ref{ss:choosingtheobjectivefunction}, SNIP and GraSP can be seen as optimizing the objectives \(\mathcal{R}_{\Delta \mathcal{L}}\) and \(\mathcal{R}_{\lVert \nabla \rVert_2}\), respectively (cf. \eqref{saliencydefinition}). Plotted values are calculated from VGG-16 on CIFAR-10 at initialization. Tickets drawn through saliency pruning rarely outperform baselines, even on the objectives they are designed to optimize.}
    \label{fig:theissuewithsaliency}
\end{figure} 

In recent years, the demand for highly-sparse subnetworks that can be easily deployed to edge devices has increased \cite{edgedevices}. In such highly-sparse regimes, however, ignoring these inter-weight interactions leads to drastic drops in performance. In fact, in the sparse regime, we find that saliency-based methods cannot even adequately optimize their own objective function \(\mathcal{R}\) relative to other methods. 

To illustrate this, we consider two popular saliency-based pruners: SNIP~\cite{snip}, defined as \(\mathcal{S}(\theta_j) = \lvert \tfrac{\partial \mathcal{L}}{\partial \theta_j} \cdot \theta_j \rvert\), and GraSP~\cite{grasp}, defined as \(\mathcal{S}(\theta_j) = -( \boldsymbol{H} \tfrac{\partial \mathcal{L}}{\partial \theta_j} )_j \cdot \theta_j\). As described in Section~\ref{ss:choosingtheobjectivefunction}, the GraSP criterion is equivalent to letting \(\mathcal{R}\) in \eqref{saliencydefinition} be the negative gradient norm, \(\mathcal{R}_{\lVert \nabla \rVert_2} := - \lVert \nabla_{\theta} \mathcal{L} \rVert_2\), and the SNIP criterion is approximately equivalent to letting it be the relative change in loss, \(\mathcal{R}_{\Delta \mathcal{L}} := \lvert \tfrac{\mathcal{L}}{\mathcal{L}_{\text{dense}}} - 1 \rvert\). 

Fig.~\ref{fig:theissuewithsaliency} compares the performance of these saliency-based methods, evaluated with their respective objective functions, against LTR~\cite{instability}, which produces lottery tickets, and naive magnitude pruning under identical initializations. We simulate this for VGG-16~\cite{vgg16} on the CIFAR-10 dataset~\cite{cifar10}. 

We note that LTR does not prune at initialization, and thus its corresponding ticket need not be strong at this stage, and especially not in regard to arbitrarily defined objective functions. Regardless, as seen in Fig.~\ref{fig:theissuewithsaliency}, the saliency-based methods perform \textit{worse} than LTR on a relatively consistent basis, particularly in the sparse regime. Furthermore, it can be surpassed by the naive magnitude pruning baseline, which does not produce lottery tickets.

Therefore, in our work, we opt to move away from the first-order approximations of synaptic saliency and towards a holistic ticket optimization approach, in which entire groups of important connections are learned simultaneously.

\section{Methodology}

Given a neural network, we aim to draw lottery tickets early in training while avoiding the shortcomings of the first-order approximations of Synaptic Saliency. To this end, we frame the problem as a constrained combinatorial optimization over the entire ticket and develop a probabilistic, differentiable search algorithm that directly enforces the desired sparsity: Concrete Ticket Search (CTS). Importantly, rather than implemented throughout training, ticket search is applied in a single shot while the network weights are frozen to capitalize on the benefits of the empirically observed stable phase of training \cite{instability}, \cite{earlyphase}, \cite{criticallearningperiods}. In the advent of hardware support for efficient sparse training, this would allow for significant training computation reductions. This also improves optimization stability; the ticket search algorithm is not interrupted by noise inherent in traditional stochastic gradient descent. 

\subsection{Problem Formulation} \label{ss:problemformulation}

\subsubsection{The discrete problem}

Suppose we are given a target density \(\kappa \in (0, 1]\), representing the desired fraction of active (unpruned) weights in some state of a neural network \(f(\cdot; \theta)\) with parameters \(\theta \in \mathbb{R}^d\) held fixed. We seek a binary mask \(m '
 \in \{0, 1\}^d\) that minimizes an objective function \(\mathcal{R}\) over a dataset \(\mathcal{D} = \{(x_i, y_i)\}_{i=1}^N\). We intentionally leave the auxiliary objective \(\mathcal{R}\) variable as it may differ from the standard task loss. The task of ticket searching is:
\begin{align}
	& \min_{m \in \{0, 1\}^d} \mathcal{R}(m; \mathcal{D}) = \min_{m \in \{0, 1\}^d} \mathbb{E}_{(x, y) \sim \mathcal{D}} \left[ \mathcal{R}(f(x; m \odot \theta), y) \right] \nonumber \\
	& \ \text{s.t.} \  \frac{\lVert m \rVert_0}{d} = \kappa \ \text{where} \ \lVert m \rVert_0 = \sum_{j=1}^d \mathbb{I}[m_j \neq 0], \label{e:problemformulation}
\end{align}
where \(\odot\) denotes the element-wise product. Here, the discrete nature of \(m\) prevents gradient-based optimization, and the search space of \(\binom{d}{\kappa d}\) possible masks is intractable for combinatorial optimization over any non-trivial \(d\).

\subsubsection{The probabilistic relaxation}
We relax the discrete constraint by parameterizing a probability distribution over the space of masks. We introduce a trainable vector of retention probabilities \(\alpha \in [0, 1]^d\) and define a distribution \(q_\alpha\) where each mask element \(m_j\) is sampled independently from a Bernoulli distribution:
\(m_j \stackrel{\text{i.i.d.}}{\sim} \textrm{Bernoulli}(\alpha_j)\). Then, \(\lVert m \rVert_0\) can be written in terms of its expectation,
\begin{equation}
    \mathbb{E}_{m \sim q_{\alpha}} [\|m\|_0] = \mathbb{E} \left[ \sum_{j=1}^{d} \mathbb{I}[m_j \neq 0] \right] = \sum_{j=1}^{d} \mathbb{E}[m_j] = \sum_{j=1}^{d} \alpha_j
\label{e:expectationl0norm}
\end{equation}
and, assuming \((x, y) \sim \mathcal{D}\) for brevity, the relaxation of \eqref{e:problemformulation} is

\begin{equation}
	\min_{\alpha \in [0,1]^d} \mathbb{E}_{m \sim q_{\alpha}} \left[ \mathcal{R} (f(x; m \odot \theta), y) \right] \ \ \text{s.t.} \ \frac{1}{d} \sum_{j = 1}^d \alpha_j  \approx \kappa. \label{e:continuousformation}
\end{equation}
In practice, we produce a deterministic mask at the end by retaining only the most probable weights, i.e., taking \(m = \textrm{top-k}(\alpha, \kappa)\) where 
\begin{equation}
    \textrm{top-k}(\alpha, \kappa)_j = \mathbb{I}[\alpha_j \geq \overline{\alpha_\kappa}],
\label{e:topkclamp}
\end{equation}
and \(\overline{\alpha_\kappa}\) is the \((\kappa d)\)-th largest element of \(\alpha\).

Now, although the optimization variables are continuous, another challenge arises. The expectation is taken over discrete samples of \(m\), which blocks the flow of gradients. To compute \(\nabla_{\alpha} \mathbb{E}_{m \sim q_\alpha}[ \cdot ]\), a gradient estimator is required.

\subsection{The Concrete Relaxation}

As our gradient estimator, we use a Concrete relaxation~\cite{concretedistribution}, which constructs a continuous and differentiable approximation of the discrete Bernoulli sampling process. Two other common approaches include score-function estimators, \cite{estimatorreinforce}, \cite{estimatorrebar}, \cite{estimatorrelax}, \cite{estimatorarm}, which use the log-derivative trick and straight-through estimators, \cite{estimatorste}, which pretend discrete operations are continuous. In practice, however, the former of these suffers from extremely high variance due to its stochastic estimation of returns. This high variance severely slows convergence and can lead to very unstable optimization dynamics. In contrast, the latter suffers from biased gradients due to ignoring non-differentiable operations. Along with the fact that these gradients do not correspond to the true objective, they unfairly stagnate many weights in the optimization space and can lead to situations where many weights are never even considered by the optimization algorithm. Again, this severely slows the drawing of an optimal ticket.

The Gumbel-softmax reparameterization addresses these limitations by sacrificing the use of exact discrete samples during optimization. Instead, the retention probabilities (i.e., \(\alpha\)) are sampled into a soft approximation of a mask, which remains concentrated around the values of 0 and 1. This allows for gradients to be exactly propagated to the retention probabilities, resulting in significantly lower variance and unbiased gradients, unlike the other methods. In addition, it avoids weight stagnation by allowing all weights to have meaningful gradients through optimization, which expedites the process.  

\subsubsection{The Gumbel-softmax reparameterization}
This process starts with the Gumbel-max trick, which allows for sampling from a categorical distribution. For a Bernoulli trial, it states that we can express the sampling of \(m_j \sim \text{Bernoulli}(\alpha_j)\) as:
\begin{equation}
    m_j = \underset{i \in \{ 0, 1 \}}{\arg\max} \left( \log\left(\mathbb{P}[m_j = i]\right) + g_i \right),
\label{e:gumbelmax}
\end{equation}
where \(g_0, g_1 \sim \textrm{Gumbel}(0,1)\). This allows us to represent \(m\) as a deterministic function of \(\alpha\) and sampled noise \(g \sim \textrm{Gumbel}\), but the \(\arg\max\) remains non-differentiable. The Gumbel-softmax trick replaces it with a softmax function, creating a continuous random variable \(s \in [0,1]^d\) that approximates the mask \(m\), representing the probability of \(m_j = 1\) in the sample:
\begin{equation}
    s_j = \tfrac{\exp\left( (\log \alpha_j + g_1) / \tau \right)}{\exp\left( (\log \alpha_j + g_1) / \tau \right) + \exp\left( (\log (1 - \alpha_j) + g_0) / \tau \right)},
\label{e:gumbelsoftmax}
\end{equation}
where \(\tau \in (0, \infty)\) is a temperature parameter. as \(\tau \rightarrow 0\), \(s_j\) approaches a Bernoulli sample. 

Additionally, since the difference of two independent Gumbel variables follows a logistic distribution, with some algebraic manipulation \eqref{e:gumbelsoftmax} can be rewritten simply as 
\begin{equation}
s_j = \sigma\left(\tfrac{\textrm{logit}(\alpha_j) + (g_1 - g_0)}{\tau}\right) = \sigma\left(\tfrac{\textrm{logit}(\alpha_j) + \varepsilon_j}{\tau}\right),
\label{e:sigmoidbinconcrete}
\end{equation}
where \(\sigma\) is the sigmoid function, \(\textrm{logit}\) is the inverse sigmoid function, and \(\varepsilon_j \sim \textrm{logistic}\). For numerical stability and convenience of \eqref{e:sigmoidbinconcrete} we store and optimize the logit-parameters \(\alpha_{\textrm{logit}} := \textrm{logit}(\alpha)\). This improves conditioning of gradient updates and widens the domain to \(\mathbb{R}^d\) (from \([0,1]^d\)).

Crucially, the sample \(s_j\) is now a deterministic and differentiable function of the parameter \(\alpha_j\) and an independent noise source \(\varepsilon_j\). The randomness is now externalized and fixed with respect to \(\alpha\), allowing us to move the gradient operator inside the expectation. As such, the gradient of our objective function with respect to \(\alpha\) can be approximated by differentiating through the soft mask \(s\):
\begin{equation}
\begin{split}    
\nabla_\alpha \mathbb{E}_{m \sim q_\alpha}&\left[ \mathcal{R} \right] \approx \nabla_\alpha \mathbb{E}_{s \sim \text{Concrete}(\alpha)}\left[ \mathcal{R}(f(x; s \odot \theta), y) \right] \\
&= \mathbb{E}_{\varepsilon \sim \text{logistic}} \left[ \nabla_\alpha \mathcal{R} (f(x; s(\alpha, \varepsilon) \odot \theta), y) \right].
\end{split}
\label{e:gradientreparameterized}
\end{equation}
This reparameterization thus allows for the use of standard Monte Carlo gradient descent methods under a low-variance and only slightly biased gradient.

\subsubsection{Sparsity control}

\begin{figure}[!t] 
%\caption{Gradient Update Algorithms}
\begin{algorithm}[H]
\caption{\textsc{Lagrange}: Gradient Step} \label{lagrangemultiplieralgoirthm}
\begin{algorithmic}[1]
    \Require Frozen network parameters \(\boldsymbol{\theta}\), logit parameters \(\boldsymbol{\alpha_\textrm{logit}}\), lagrange multiplier \(\lambda\),  target density \(\kappa\), concrete temperature \(\tau\), objective function \(\mathcal{R}\), data batch \(\mathcal{D}_b = \{ (x_i, y_i) \}^b_{i = 1}\)
    \Ensure Gradients \((\boldsymbol{g_\alpha}, g_\lambda)\) for parameters \(\boldsymbol{\alpha_\textrm{logit}}\) and \(\lambda\).
    \State \(\boldsymbol{\varepsilon} \stackrel{\text{i.i.d.}}{\sim} \text{logistic}(0, 1)\)
    \State \(\boldsymbol{s} \gets \sigma ((\boldsymbol{\alpha_\textrm{logit}} + \boldsymbol{\varepsilon})/\tau)\) \Comment{cf. \eqref{e:sigmoidbinconcrete}}
    \State \(\boldsymbol{\theta_{\text{eff}}} \gets \boldsymbol{s} \odot \boldsymbol{\theta}\)
    \State \(\mathcal{L}_{\text{sparsity}} \gets \operatorname{sum} (\sigma(\boldsymbol{\alpha_\textrm{logit}})) \cdot \tfrac{1}{\kappa d}- 1\) \Comment{cf. \eqref{e:sparsityconstraint}}
    \State \(\mathcal{R}_{b} \gets \frac{1}{b} \sum_{i = 1}^b \left(\mathcal{R}(f(x_i; \boldsymbol{\theta_{\text{eff}}}), y_i)\right)\)
    \State \(\mathcal{L} \gets  \mathcal{R}_b + \lambda \cdot \mathcal{L}_{\text{sparsity}}\)
    \State \(\boldsymbol{g_\alpha} \gets \operatorname{grad}(\mathcal{L}, \boldsymbol{\alpha_\textrm{logit}})\) 
    \State \(g_\lambda \gets -1 \cdot \operatorname{grad}(\mathcal{L}, \lambda)\)
\end{algorithmic}
\end{algorithm}

\begin{algorithm}[H]
\caption{\textsc{GradBalance}: Gradient Step} \label{gradbalancealgorithm}
\begin{algorithmic}[1]
    \Require Frozen network parameters \(\boldsymbol{\theta}\), logit parameters \(\boldsymbol{\alpha_\textrm{logit}}\), gradient balancer \(\lambda\), target density \(\kappa\), concrete temperature \(\tau\), smoothing factor \(\eta\), objective function \(\mathcal{R}\), data batch \(\mathcal{D}_b = \{ (x_i, y_i) \}^b_{i = 1}\)
    \Ensure Gradient \(\boldsymbol{g_\alpha}\) for parameters \(\boldsymbol{\alpha_\textrm{logit}}\) and updated \(\lambda\).
    \State Repeat steps 1-5 from Algorithm~\ref{lagrangemultiplieralgoirthm} for \(\mathcal{R}_b\) and \(\mathcal{L}_{\text{sparsity}}\).
    \State \(\boldsymbol{g_{\mathcal{\text{objective}}}} \gets \operatorname{grad}(\mathcal{R}_b, \boldsymbol{\alpha_{\textrm{logit}}})\)
    \State \(\boldsymbol{g_{\mathcal{\text{sparsity}}}} \gets \operatorname{grad}(\mathcal{L}_{\text{sparsity}}, \boldsymbol{\alpha_{\textrm{logit}}})\)
    \If{\(\mathcal{L}_\text{sparsity} > 0\)}
        \State \(\lambda_{\text{target}} \gets \frac{\lVert \boldsymbol{g_{\mathcal{\text{objective}}}} \rVert_{2}}{\lVert \boldsymbol{g_{\mathcal{\text{sparsity}}}} \rVert_{2}}\)
    \Else
        \State \(\lambda_{\text{target}} \gets 0\)
    \EndIf
    \State \(\lambda = \eta \cdot \lambda + (1-\eta) \cdot \lambda_{\text{target}}\)
    \State \(\boldsymbol{g_\alpha} \gets \boldsymbol{g_{\mathcal{\text{objective}}}} + \lambda \cdot \boldsymbol{g_{\mathcal{\text{sparsity}}}}\) 
\end{algorithmic}
\end{algorithm}
\label{fig:gradientupdatealgorithms}
\end{figure}

The optimization in \eqref{e:continuousformation} requires control over the expected number of active weights throughout training. The standard trick is to add an \(L_p\) regularization term (typically \(p \in \{ 1, 2\}\)) \cite{learningl0sparse}, \cite{proxsgd}, over the mask probabilities \(\alpha\), yielding minimization loss 
\[\mathcal{L} := \mathcal{R} + \lambda \mathcal{L}_{\text{reg}},\]
where \(\lambda\) is a tunable hyperparameter. While simple, this indirect scheme does not guarantee \(\mathbb{E}[\lVert m \rVert] \approx \kappa d\) and requires laborious tuning of \(\lambda\). 

Instead, we enforce the sparsity target explicitly by first defining the normalized sparsity constraint 
\begin{equation}
\mathcal{L}_{\text{sparsity}} := \frac{\mathbb{E}[\lVert m \rVert_0]}{\kappa d} - 1. \label{e:sparsityconstraint}
\end{equation}
Unlike the indirect approach, this inherently adjusts its regularization strength relative to different target densities \(\kappa\) and approaches 0 as the desired sparsity is met. We further note that we \textit{avoid} optimizing \(\lvert \mathcal{L}_{\text{sparsity}} \rvert\) to retain stability of optimization dynamics. We now introduce two practical methods, whose pseudo-code can be viewed in Algorithms~\ref{lagrangemultiplieralgoirthm}-\ref{gradbalancealgorithm}, to enforce this constraint under the optimization problem
\begin{equation}
    \min_{\alpha \in [0,1]^d} \mathbb{E}_{s \sim  \text{Concrete}(\alpha)} \left[ \mathcal{R}(f(x; s \odot \theta, y) \right]  \ \text{s.t.} \ \mathcal{L}_\text{sparsity} = 0. 
\label{e:sparsityconstraintoptimization}
\end{equation}

Perhaps the most common approach to solving constrained optimization problems is the use of Lagrange multipliers. This consists of searching for saddle points of the Lagrangian \(\mathcal{L}(\alpha, \lambda) := \mathcal{R} + \lambda \cdot \mathcal{L}_\text{sparsity}\), where \(\lambda \in \mathbb{R}\) is the Lagrange multiplier. This can thus be written as the dual optimization 
\begin{equation}
    \max_{\lambda \in \mathbb{R}} \min_{\alpha \in [0,1]^d} [\mathcal{R} + \lambda * \mathcal{L}_{\text{sparsity}}],
\label{e:lagrangemultiplieroptimization}
\end{equation}
which can be solved under traditional gradient descent methods, utilizing \eqref{e:gradientreparameterized}. Because \(\mathcal{L}_{\text{sparsity}}\) is not propagated through the network, its gradient can be computed analytically, thereby avoiding any additional overhead from automatic differentiation.

The problem with this naive dual optimization is that it has a tendency to be unstable in practice, often due to scaling differences and unconstrained interactions between gradient updates of the constraint and objective. This often necessitates careful and task-specific tuning of optimization hyperparameters, making it cumbersome in practice.

To overcome these issues, we introduce \textsc{GradBalance}, a more stable and adaptive alternative roughly inspired by GradNorm~\cite{gradnorm}. Rather than introducing any new optimization variables, \textsc{GradBalance} balances the gradients of the objective and constraint directly at each step, equalizing the influence of the objective and the constraint on the total gradient. This is achieved by adjusting \(\lambda\), now denoted the `gradient balancer', so that the magnitude of the constraint gradient matches that of the objective gradient. At each step, the target gradient balancer, is thus calculated as the ratio of the gradient norms \(\lambda_{\text{target}} := \lVert \nabla_{\alpha} \mathcal{R} \rVert_{2} / \lVert \nabla_{\alpha} \mathcal{L}_{\text{sparsity}} \rVert_{2} \). If the desired sparsity is met, \(\lambda_{\text{target}}\) is instead set to 0, removing the constraint's influence. To prevent oscillations and improve stability, the active \(\lambda\) is smoothed towards \(\lambda_{\text{target}}\) using an exponential moving average. In addition, in our implementation, we set \(\kappa_{\text{eff}} := 1.1 \kappa\) slightly larger than the input \(\kappa\) so that the sparsity constraint is not enforced too rigidly.

A notable characteristic of \textsc{GradBalance} is its one-sided application of the sparsity constraint. If the network becomes sparser than the target (i.e., \(\mathcal{L}_{\text{sparsity}} \leq 0\)), there is no explicit corrective pressure to increase retention probabilities; such updates are driven solely by the objective gradient. In the dense regime (i.e., for large \(\kappa\)), this can lead to overly sparse tickets being drawn. Nevertheless, the design of CTS is primarily based in the high-sparsity regime. Also, we find that \textsc{GradBalance} consistently provides smoother and more reliable convergence than a traditional Lagrange multiplier formulation across all sparsities, even after hyperparameter tuning of the latter. The absence of this hyperparameter tuning additionally allows for quick and robust application to varying tasks.

For both approaches, to avoid large bias due to the multiplicative nature of the soft mask, we initialize \(\alpha\) so that \(\mathcal{L}_{\text{sparsity}}\) begins at zero and all weights have an equal chance of being selected. This is achieved simply by initializing \(\alpha = \kappa \cdot \boldsymbol{1}_d\).

\subsection{Choosing the Objective Function} \label{ss:choosingtheobjectivefunction}

In the design for CTS, \(\mathcal{R}\) can be any differentiable objective function. In our experiments. we investigate several choices for \(\mathcal{R}\), drawing inspiration from existing pruning criteria as well as newer principles of network stability.

A straightforward choice for the objective function is the standard task loss, \(\mathcal{R}_{\mathcal{L}} := \mathcal{L}\). Although there is no guarantee that tickets exhibiting low task loss near initialization are genuine lottery tickets, a ticket's initial performance provides a lower bound for accuracy. The pruning criterion for SNIP~ \cite{snip} is closely related: \(\mathcal{S}(\theta_j) = \lvert \frac{\partial \mathcal{L}}{\partial \theta_j} \cdot \theta_j \rvert\). However, due to its focus on saliency magnitude, this score prioritizes weights that cause large changes in the loss, rather than those that strictly minimize it. To capture this objective directly, we define an objective based on the relative change in loss, \(\mathcal{R}_{\Delta \mathcal{L}} := \lvert \tfrac{\mathcal{L}}{\mathcal{L}_{\text{dense}}} - 1 \rvert\), where \(\mathcal{L}_{\text{dense}}\) is the task loss of the unpruned network \(f_\theta := f(\cdot; \theta)\). In a similar vein, the GraSP criterion~\cite{grasp}, \(\mathcal{S}(\theta_j) = - (\boldsymbol{H} \tfrac{\partial \mathcal{L}}{\partial \theta_j})_j \cdot \theta_j\), aims to preserve gradient flow through the network, inspired by works studying the neural tangent kernel \cite{neuraltangentkernel}, \cite{evci_gradient_flow}. This can be framed as maximizing the gradient norm, \(\mathcal{R}_{-\lVert \nabla_{\theta} \mathcal{L} \rVert_2} := - \lVert \nabla_{\theta} \mathcal{L} \rVert_2\).

In addition to these, we propose three novel objectives motivated by the training dynamics of neural networks. Recent works indicate that networks undergo an unstable ``early phase'' \cite{earlyphase}, \cite{butterflyeffect}, before settling into a stable training trajectory; they are sensitive to perturbations of data \cite{criticallearningperiods}, the Hessian Eigenspectrum is diverse \cite{hessianeigenvalue}, and subsequent training runs converge to different minima in state space (i.e., convergence weights are not linearly mode connected) \cite{instability}. During this period, IMP is ineffective, which is what prompted the rise of rewinding to an iteration near the beginning of training rather than at initialization \cite{instability}. Furthermore, it has been shown that, unlike failing methods, lottery tickets drawn through LTR themselves exhibit high stability during training and follow similar learning trajectories to their dense counterpart, \cite{evci_gradient_flow}, \cite{earlyphase}, \cite{instability}. We thus attempt to preserve training dynamics by drawing inspiration from knowledge distillation, treating the dense network \(f_\theta\) as the teacher and the ticket \(f_{s \odot \theta}\) as the student. In these cases, it is necessary to run two forward passes at each gradient step: one for the teacher (which is not contained in the gradient's computation graph) and one for the student. Regardless, in comparison to the colossal amount of computation required for LTR, we view this as a viable concession.

One standard method of knowledge distillation is to directly ensure that the student produces a similar output distribution as the teacher. As such, we minimize the reverse KL divergence of their output distributions as in \cite{minillm}, \cite{rethinkingkld}, \cite{selfdistillationdropout}: 
\[\mathcal{R}_{\text{KL}} := D_{\text{KL}}\left(\textrm{softmax}(f_{s \odot \theta}(\cdot)) \parallel \textrm{softmax}(f_\theta (\cdot)) \right). \]
Additionally, to preserve the internal computations of the network, we encourage feature maps of the sparse network to align with those of the dense counterpart. We measure the dissimilarity using Mean Squared Error (MSE) between normalized layer-wise feature activations, similar to \cite{fitnets}, \cite{angulardistillation}. Let \(F^{(l)}_{\text{ticket}}\) and \(F^{(l)}_{\text{parent}}\) be the feature activations at layer \(l\). Then, letting \(L\) be the number of layers, we define
\[\mathcal{R}_{\text{feature}} := \frac{1}{L} \sum_{i = 1}^{L} \operatorname{MSE} \left( \mathcal{N}(F^{(l)}_{\text{ticket}}), \mathcal{N}(F^{(l)}_{\text{parent}})  \right),\]
where \(\mathcal{N}(X) = (X - \mathbb{E}(X))/\sqrt{\operatorname{Var}(X)}\) is a per-layer normalization function. Stepping away from knowledge distillation, to most directly enforce a similar learning trajectory, we match the gradients of the sparse network's parameters to those of the dense network. A subnetwork that has its first gradient step in the same direction as its parent is more likely to converge to a similar solution. This objective is formulated similarly to feature matching:
\[\mathcal{R}_{\text{grad}} := \frac{1}{L} \sum_{i = 1}^{L} \operatorname{MSE} \left( \mathcal{N}(\nabla_{\theta^{(l)}} \mathcal{L}), \mathcal{N}(\nabla_{\theta^{(l)}} \mathcal{L}_{\text{dense}})  \right),\]
where \(\theta^{(l)}\) correspond to the parameters of layer \(l\).

\subsection{Further Discussion} \label{ss:methodologydiscussion}

\begin{figure}[!t] 
%\caption{Algorithm for Generating Subnetworks}
\begin{algorithm}[H]
\caption{\(\textsc{CTS}_{\mathcal{R}}\)}
\begin{algorithmic}[1]
    \Require Pruning density \(\kappa\), network \(f\) with original parameters \(\boldsymbol{\theta}_0\), training data \(\mathcal{D}\), ticket search steps \(S\), rewinding iteration \(k\), training steps \(T\).
    \Ensure Convergence weights of drawn ticket \(\boldsymbol{\theta}_{f}\), with density \(\kappa\).     
    \State Train the network \(f_{\boldsymbol{\theta}}\) on \(\mathcal{D}\) for \(k\) steps to obtain \(\boldsymbol{\theta}_k\)
    \State Freeze \(\boldsymbol{\theta}_k\) and initialize \(\boldsymbol{\alpha_{\textrm{logit}}} \gets \textrm{logit}(\kappa) \cdot \boldsymbol{1}_d\), \(\lambda = 0\).
    \For{$i = 1$ to $S$}
        \State Sample minibatch $\mathcal{D}_b = \{ (x_i, y_i) \}^b_{i=1} \sim \mathcal{D}$.
        \iffalse
        \If{using Algorithm~\ref{lagrangemultiplieralgoirthm}}
            \State $(\boldsymbol{g_\alpha}, g_\lambda) \gets \textsc{Lagrange}(\boldsymbol{\theta}_k, \boldsymbol{\alpha_{\textrm{logit}}}, \lambda, \mathcal{D}_b)$
            \State Update $\boldsymbol{\alpha_{\textrm{logit}}} \gets \text{OptStep}(\boldsymbol{\alpha_{\textrm{logit}}}, \boldsymbol{g_\alpha})$
            \State Update $\lambda \gets \text{OptStep}(\lambda, g_\lambda)$
        \ElsIf{using Algorithm~\ref{gradbalancealgorithm}}
        \fi
            \State $(\boldsymbol{g_\alpha}, \lambda) \gets \textsc{GradBalance}(\boldsymbol{\theta}_k, \boldsymbol{\alpha_{\textrm{logit}}}, \lambda, \mathcal{D}_b)$
            \State Update $\boldsymbol{\alpha_{\textrm{logit}}} \gets \text{OptStep}(\boldsymbol{\alpha_{\textrm{logit}}}, \boldsymbol{g_\alpha})$
        %\EndIf
    \EndFor
    \State \(\boldsymbol{m} \gets  \textrm{top-k}(\boldsymbol{\alpha_{\textrm{logit}}} , \kappa)\) \Comment{cf. \eqref{e:topkclamp}}
    \State Train \(f_{\boldsymbol{m} \odot \boldsymbol{\theta}}\) on \(\mathcal{D}\) for \(T-k\) steps to obtain \(\boldsymbol{\theta}_{f}\).
\end{algorithmic}
\label{subnetworkgenerationalgorithm}
\end{algorithm}
\end{figure}

So far in this section, we have illustrated the ticket search algorithm, but have not given details on how to apply it. The complete procedure for finding and training a subnetwork using CTS with \textsc{GradBalance} is outlined in Algorithm~\ref{subnetworkgenerationalgorithm}. Using the \textsc{Lagrange} gradient step is identical aside from step 5. Motivated by \cite{instability}, \cite{earlyphase}, and the design of our metrics that preserve training dynamics, we wait for a brief initial training period of \(k\) steps before performing ticket search. Then, after applying the ticket search algorithm to draw out ticket, we continue training until convergence. 

As in \cite{learningl0sparse} and the appendix of \cite{concretedistribution}, we opt to not anneal the concrete temperature \(\tau\). This is meant to preserve optimization stability and not interfere with the dynamical gradient strength in \textsc{GradBalance}. We also note that we avoid data augmentation during the ticket search phase. This tends to be significantly faster, and the benefits of data augmentation are not demanded during the search. Since the task involves selecting subnetworks over fixed weights, as opposed to tuning the weights themselves, it is more constrained, and overfitting, the primary impetus for data augmentation, is not a concern.

\section{Experiments}

In this section, we present experimental results of CTS under different objective functions. We include results on VGG-16~\cite{vgg16} and ResNet-20~\cite{resnet20} on CIFAR-10~\cite{cifar10}, and on ResNet-50~\cite{resnet20} on ImageNet~\cite{imagenet}. We first denote experimental settings and, on CIFAR-10 tasks, discuss the performance of the proposed methods in the sparsity-accuracy tradeoff. Then, we show that our method passes sanity checks and discuss aspects of the drawn tickets. Next, we compare the performance of CTS with other methods under different objective functions, and illustrate the gap between our holistic ticket search compared to saliency-based methods. Finally, we conclude by displaying accuracy results on ResNet-50 on ImageNet for select sparsities, along with wall-clock time comparisons of LTR vs CTS.

\begin{figure*}
    \centering
    \includegraphics[width=\textwidth]{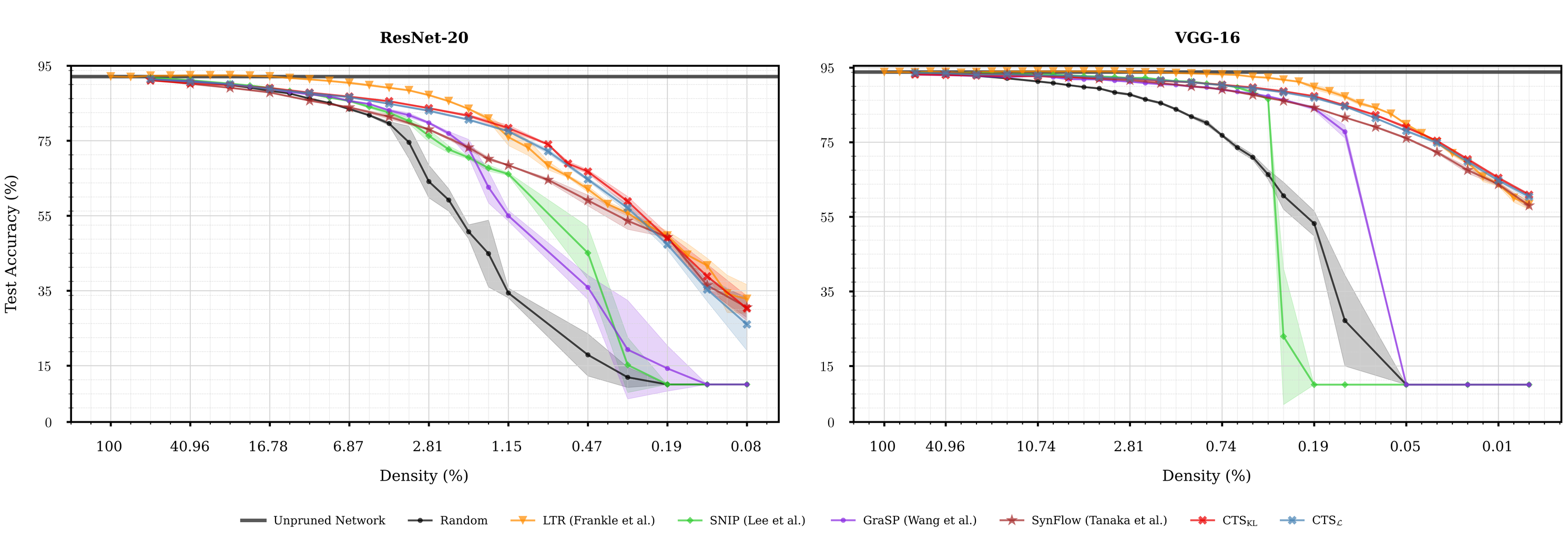}
    \caption{Test-Accuracy with respect to sparsity of subnetworks of ResNet-20 (left) and VGG-16 (right) on CIFAR-10 produced by the proposed method, LTR, SNIP, GraSP, and SynFlow. Shaded intervals are confidence intervals, taken over three runs. We plot two objectives of CTS, \(\mathcal{R}_{\text{KL}}\) and \(\mathcal{R}_{\mathcal{L}}\). Subnetworks for CTS are generated according to Algorithm~\ref{subnetworkgenerationalgorithm}, with the \textsc{GradBalance} gradient step. }
    \label{fig:accuracysparsitymain}
\end{figure*}

\begin{figure}
    \centering
    \includegraphics[width=\linewidth]{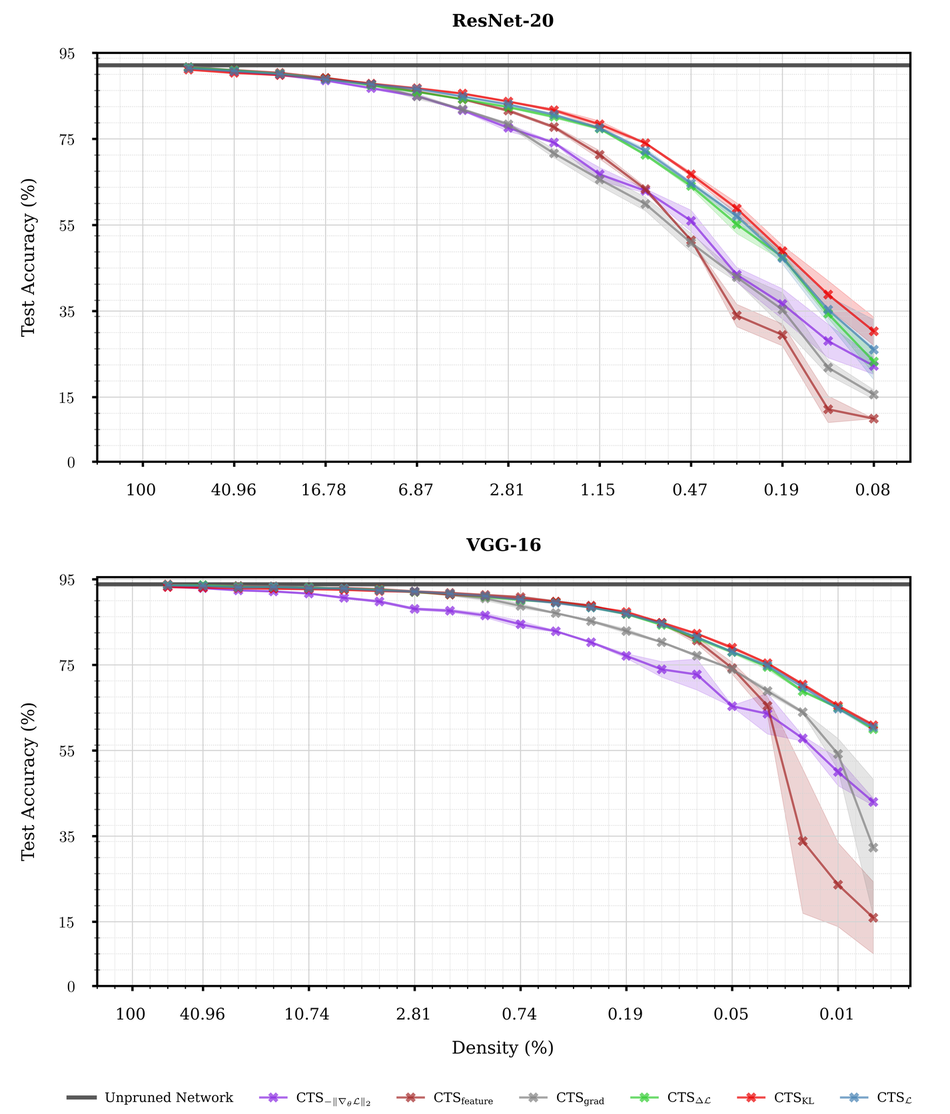}
    \caption{Test-Accuracy with respect to sparsity of subnetworks of ResNet-20 (top) and VGG-16 (bottom) on CIFAR-10 produced by the proposed method over the six objective functions outlined in Section~\ref{ss:choosingtheobjectivefunction}. Shaded intervals are confidence intervals, taken over three runs. Subnetworks for CTS are generated according to Algorithm~\ref{subnetworkgenerationalgorithm}, with the \textsc{GradBalance} gradient step. Tickets drawn under the \(\mathcal{R}_{-\lVert \nabla_{\theta} \mathcal{L} \rVert_2}\), \(\mathcal{R}_{\text{feature}}\), and \(\mathcal{R}_{\text{grad}}\) objectives cannot match performance of those drawn under \(\mathcal{R}_{\text{KL}}\), \(\mathcal{R}_{\Delta \mathcal{L}}\), and \(\mathcal{R}_{\mathcal{L}}\). }
    \label{fig:accuracysparsitycts}
\end{figure}

\subsection{Experimental Configuration} \label{ss:experimentalconfig} 
We test our holistic ticket search algorithm on CIFAR-10~\cite{cifar10} with model architectures ResNet-20~\cite{resnet20} and VGG-16~\cite{vgg16} (with batch normalization and no feedforward section) and on ImageNet~\cite{imagenet} with model architecture ResNet-50~\cite{resnet20}. As suggested by \cite{concretedistribution}, we set \(\tau = 2/3\) for all experiments. All experiments were conducted on four NVIDIA RTX Ada 6000 GPUs, four NVIDIA L40S GPUs, or four NVIDIA A30 GPUs. All results are averaged over three randomly seeded runs. 

\subsubsection{CIFAR-10 Tasks} For CIFAR-10, we follow the typical training hyperparameter configuration, e.g., \cite{lotterytickethypothesis}, \cite{snip}, \cite{grasp}, \cite{synflow}. We augment data by randomly flipping horizontally, and randomly shifting by up to four pixels in any direction, and normalizing per-channel. For both models, we train for 160 epochs using SGD with \(0.9\) momentum and weight decay of $10^{-3}$. We use a starting learning rate of $0.1$ with $10$ times drops at epochs $80$ and $120$.  To effectively utilize the hardware, we used a batch size of 512 (though rewinding iterations have been normalized to the standard of 128). For LTR, we follow the standard rate of pruning \(20\%\) per iteration, \cite{lotterytickethypothesis}, \cite{instability}, \cite{learningraterewinding}.

\subsubsection{ImageNet} For ResNet-50 on ImageNet, we follow the training hyperparameter configuration of \cite{instability} and the Github repository open\_lth~\cite{openlth}. Data is augmented by randomly flipping horizontally, randomly stretching and cropping to 224 by 224 dimension, and normalizing per-channel. We train for $90$ epochs using SGD with $0.9$ momentum and weight decay of $10^{-4}$. We use a starting learning rate of $0.4$, with linear warmup for $5$ epochs, and with $10$ times drops at epochs $30$, $60$, and $80$. We use a batch size of $1024$. Because of its extreme computational demand, most works do not include LTR as a baseline for ResNet-50 on ImageNet. To make LTR computationally viable, we prune \(53.58\%\) at each iteration, so that every 3 iterations we prune by $90\%$. Pruning by $90\%$ corresponds to approximately 10 iterations of the traditional \(20\%\) per iteration prune rate of smaller models. 

As supported in the graphs of \protect\cite{instability}, we set the rewinding iterations for LTR and the proposed method to $k=3000$ (7.5 epochs) for ResNet-20, $k=6000$ (15 epochs) for VGG-16, and $18$ epochs for ResNet-50. This is close to the range where benefits from rewinding saturate, which is necessary for the highly sparse regimes we consider.

For ticket search, i.e., learning \(\alpha\), we use the Adam optimizer without weight decay, with a learning rate of $0.1$ and a $10$ times drop at \(90\%\) of total ticket search epochs. The use of \textsc{GradBalance} stabilizes optimization itself, omitting the traditional need for low learning rates with Adam. We use an equal number of ticket search epochs as training, i.e., 160 epochs for both ResNet-20 and VGG-16. We also denote Quick CTS, which uses \(1/8\) of the time to train, i.e., 20 epochs for VGG-16 and ResNet-20. Because of the computational load of ResNet-50 on ImageNet, we only use half of the training epochs during ticket search (consequently denoted Half CTS). 

We note that our implementation of batch normalization does not track running stats (i.e., each batch is normalized with its own statistics), though baseline accuracies (including PaI and LTR methods) remain consistent with other works.

\subsection{Sparsities} Though we plot results over a wide range of sparsities, we largely focus on the highly-sparse regime (i.e., \(>50\) times compression on ResNet-20 and \(>500\) times compression on VGG-16). We make this choice because this is the region in which the issue with saliency is distinctly prominent. For ResNet-50 on ImageNet, we present results for \(10\), \(100\), and \(1000\) times compression.

\subsection{Sparsity-Accuracy Comparison on CIFAR-10}
In Fig.~\ref{fig:accuracysparsitymain}, we plot performance results for the proposed method under two selected objective functions, \(\mathcal{R}_{\text{KL}}\) and \(\mathcal{R}_{\mathcal{L}}\). As introduced previously, we use baselines of LTR~\cite{instability}, SNIP~\cite{snip}, GraSP~\cite{grasp}, and SynFlow~\cite{synflow}, along with random pruning (applied at initialization). CTS outperforms LTR in the sparse regime (particularly on ResNet-20) and remains comparable throughout. In the denser regimes, the shortcomings of saliency-based methods are less prominent.

In Fig.~\ref{fig:accuracysparsitycts}, we plot the sparsity-accuracy tradeoff for all six pruning objectives outlined in Section~\ref{ss:choosingtheobjectivefunction}. We find that maximizing the gradient norm, as proposed in \cite{grasp}, is a particularly weak method. As we display in Section~\ref{ss:objectiveperformance}, our algorithm is able to maximize gradient norm spectacularly; its weak performance points to a flaw in the paradigm of preserving gradient norm itself. We also note that both objectives that involve normalized distances (i.e., \(\mathcal{R}_{\text{feature}}\) and \(\mathcal{R}_{\text{grad}}\)) at each layer perform poorly. This is likely due to the excessive noise present as the model enters the sparse regime.

\begin{figure}
    \centering
    \includegraphics[width=0.9\linewidth]{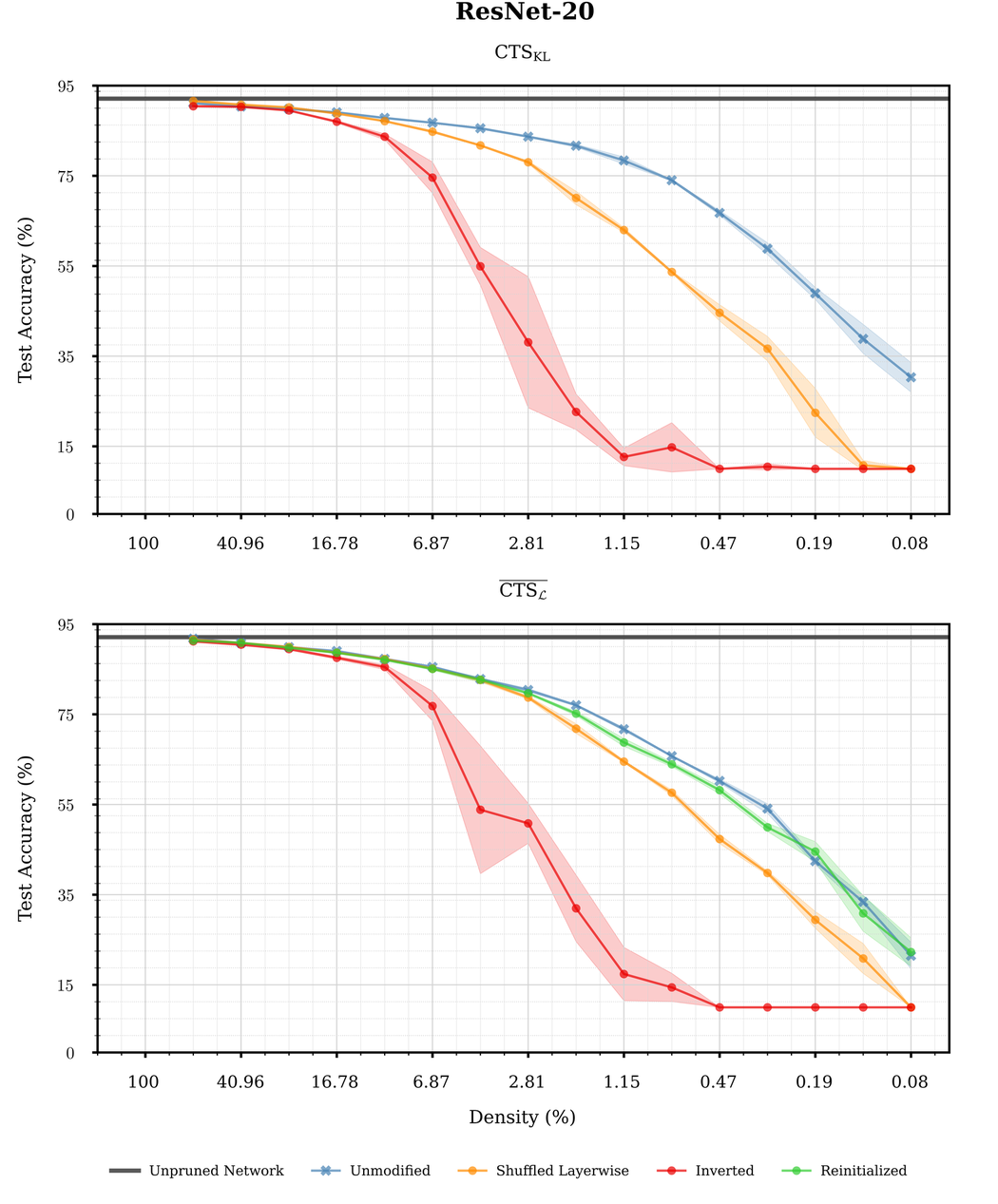}
    \caption{Test-Accuracy with respect to sparsity of subnetworks of ResNet-20 on CIFAR10, under sanity check methods suggested in \cite{pruningatinitialization}. These include weight reinitialization, score inversion, and shuffling layerwise. We test these for CTS\(_{\text{KL}}\) (top) and \(\overline{\text{CTS}_{\mathcal{L}}}\) (bottom), (cf. Section~\ref{sanitycheckingsection}). Shaded intervals are confidence intervals, taken over three runs. Tickets drawn through CTS consistently outperform those with sanity-checking ablations, especially in the sparse regime.}
    \label{fig:sanitychecksresnet}
\end{figure}

\begin{figure}
    \centering
    \includegraphics[width=0.9\linewidth]{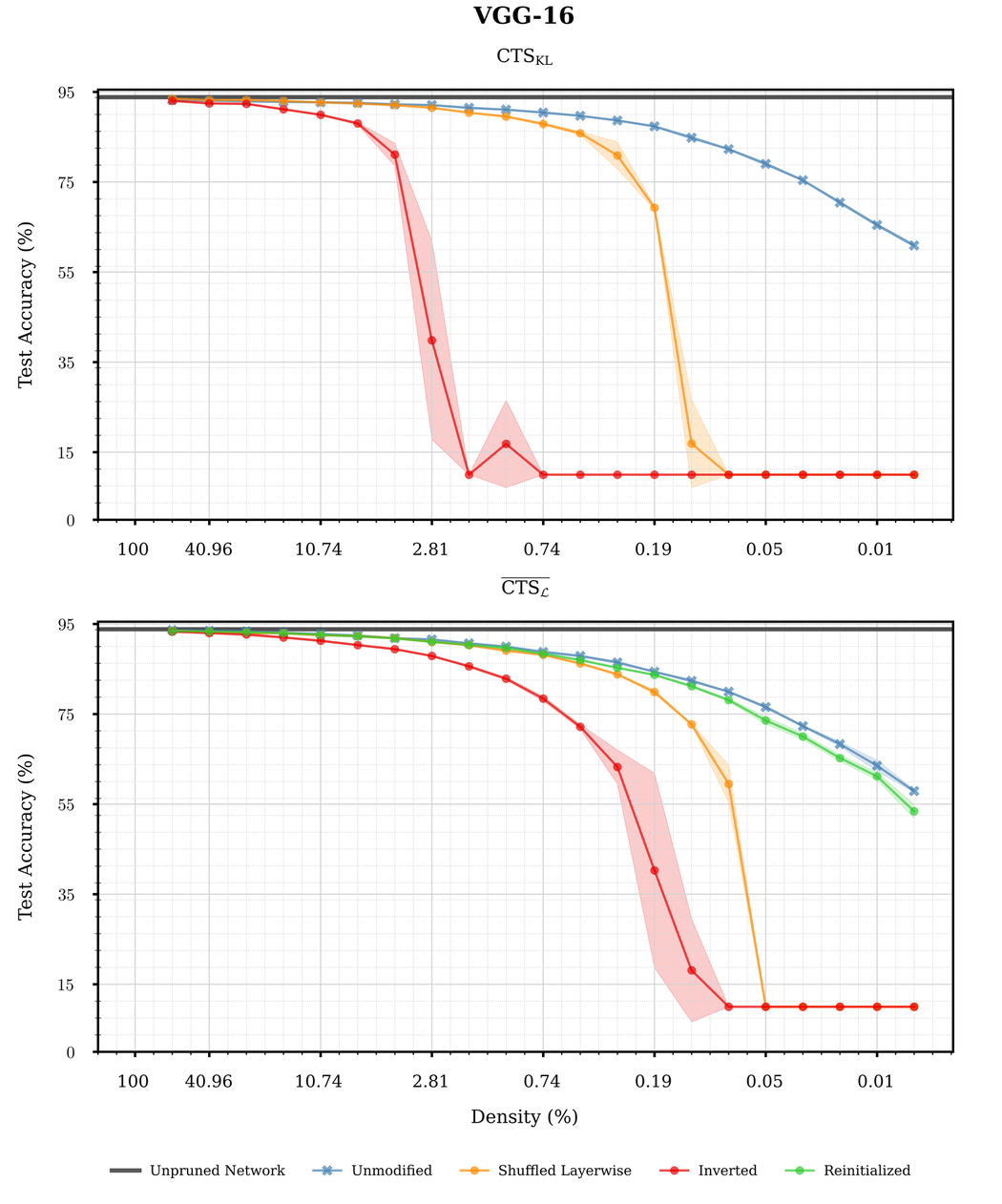}
    \caption{Test-Accuracy with respect to sparsity of subnetworks of VGG-16 on CIFAR10, under sanity check methods suggested in \cite{pruningatinitialization}. These include weight reinitialization, score inversion, and shuffling layerwise. We test these for CTS\(_{\text{KL}}\) (top) and \(\overline{\text{CTS}_{\mathcal{L}}}\) (bottom), (cf. Section~\ref{sanitycheckingsection}). Shaded intervals are confidence intervals, taken over three runs. Tickets drawn through CTS consistently outperform those with sanity-checking ablations, especially in the sparse regime.}
    \label{fig:sanitychecksvgg}
\end{figure}

\begin{figure}
    \centering
    \includegraphics[width=\linewidth]{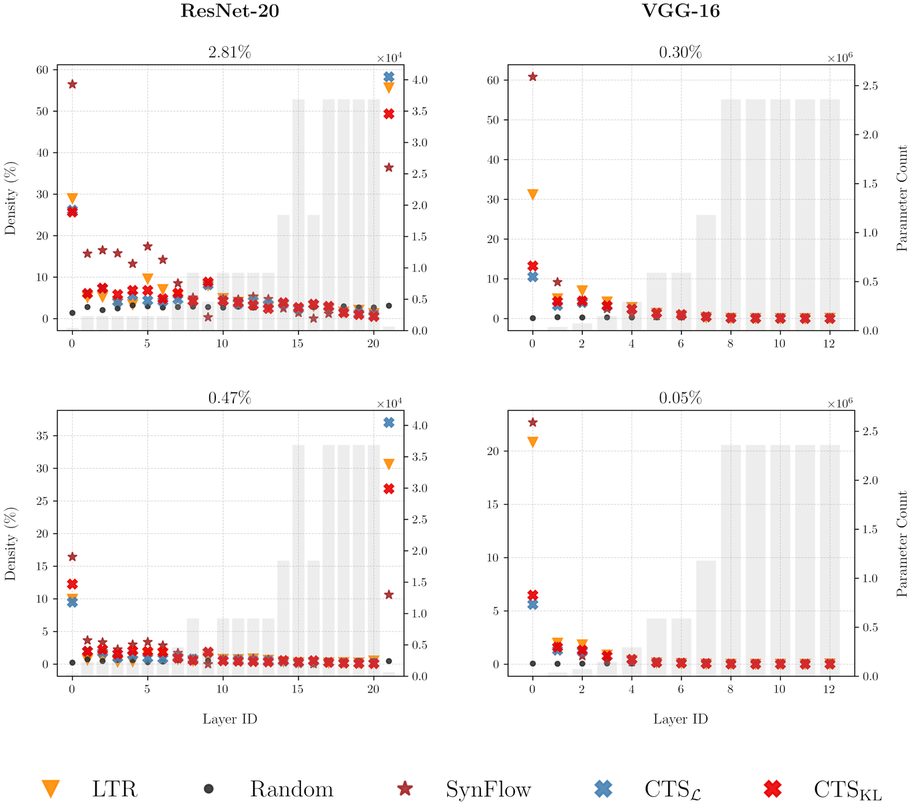}
    \caption{Layerwise sparsity patterns of drawn tickets on CIFAR-10 on ResNet-20 (left) and VGG-16 (right) at given sparsities. Methods shown include random pruning, LTR~\cite{instability}, SynFlow~\cite{synflow}, and CTS\(_{\text{KL}}\) and CTS\(_{\mathcal{L}}\). Relative parameter counts at each layer are shown with grey bars. On VGG-16, the final feature extractor forcibly remains dense for all methods and therefore is not plotted. All non-naive methods (i.e., all but random pruning) prune with relatively similar layerwise ratios and maintain a large portion of weights in the first and last layer.}
    \label{fig:layerwisesparsities}
\end{figure}

\subsection{Sanity Checking on CIFAR-10} \label{sanitycheckingsection}
We additionally run sanity checks, as proposed in \cite{pruningatinitialization} for methods that prune at initialization. This consists of drawing the initial mask, applying ablations, and then continuing to train to convergence. Ablations include random shuffling of the mask in each layer, weight reinitialization, and
score inversion (i.e., clamping to the least-probable weights rather than the most probable, cf. \eqref{e:topkclamp}). A successful pruning method should be sensitive to these ablations, since its mask should extract information from the interplay of individual weights. 

Since these sanity checks are designed for methods that prune at initialization and since weight reinitialization is meaningless early in training, we consider an additional pruning procedure, \(\overline{\text{CTS}_{\mathcal{L}}}\). This is equivalent to CTS\(_{\mathcal{L}}\), but does prunes at initialization by running ticket search directly on the randomly initialized weights. We opt to perform this using \(\mathcal{R}_{\mathcal{L}}\) because it is the most straightforward of the objectives to assess subnetworks that are unstable. We also test the sanity checks other than reinitialization on the vanilla (applied after initialization) CTS\(_{\text{KL}}\). 

Figs.~\ref{fig:sanitychecksresnet}-\ref{fig:sanitychecksvgg} show the results for ResNet-20 and VGG-16, respectively. In both cases, CTS comfortably surpasses the layerwise shuffling and score inversions. When considered, it also consistently outperforms reinitialization, especially in the sparse regime.

Fig.~\ref{fig:layerwisesparsities} compares the layerwise sparsity patterns of select methods, which have been shown to be the only subnetwork aspect that saliency-based methods can sufficiently optimize. All methods retain the most weights on the first and last layer (which, in VGG-16, is always dense). All methods, apart from random pruning, output relatively similar layerwise sparsity ratios. As with the sparsity checks, this suggests that the benefits of CTS over other methods come from the individual weights that it selects.

\begin{figure}
    \centering
    \includegraphics[width=\linewidth]{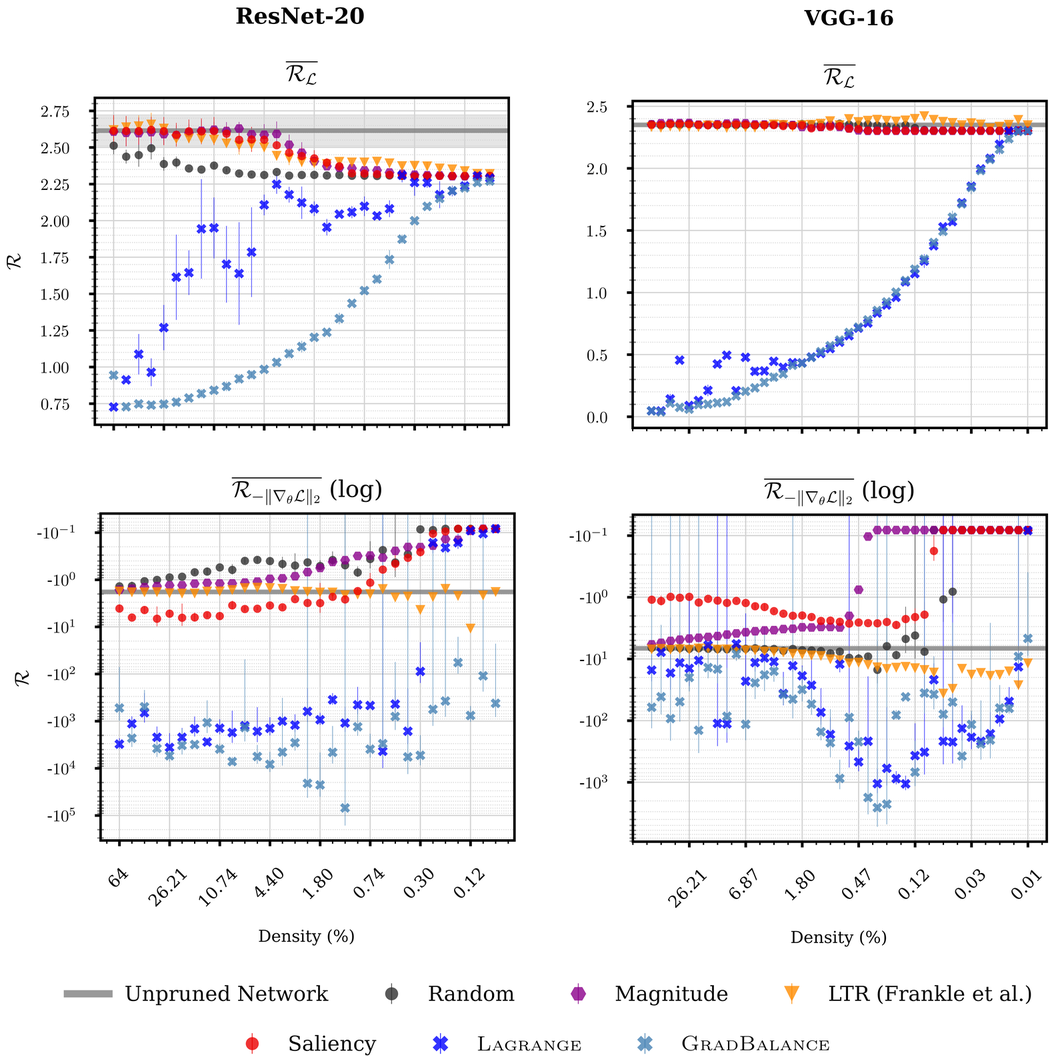}
    \caption{Objective performance with respect to sparsity of subnetworks of ResNet-20 (left) and VGG-16 (right) on CIFAR10 drawn using respective objective functions (as outlined in Section \ref{ss:choosingtheobjectivefunction}). Presented results are computed immediately after the ticket is drawn; here, we apply all methods at initialization. We plot baseline results from random pruning, magnitude pruning, LTR, and saliency pruning (i.e., SNIP for \(\mathcal{R}_{\mathcal{L}}\) and GraSP for \(\mathcal{R}_{-\lVert \nabla_{\theta} \mathcal{L} \rVert_2}\). We include results under both the \textsc{Lagrange} and \textsc{GradBalance} gradient steps, and all results are under Quick CTS. Both versions of Quick CTS overwhelmingly excel on both objectives, with \textsc{GradBalance}-based tickets performing more consistently than \textsc{Lagrange}-based ones.}
    \label{fig:objectivesinit}
\end{figure}

\begin{figure*}
    \centering
    \includegraphics[width=\textwidth]{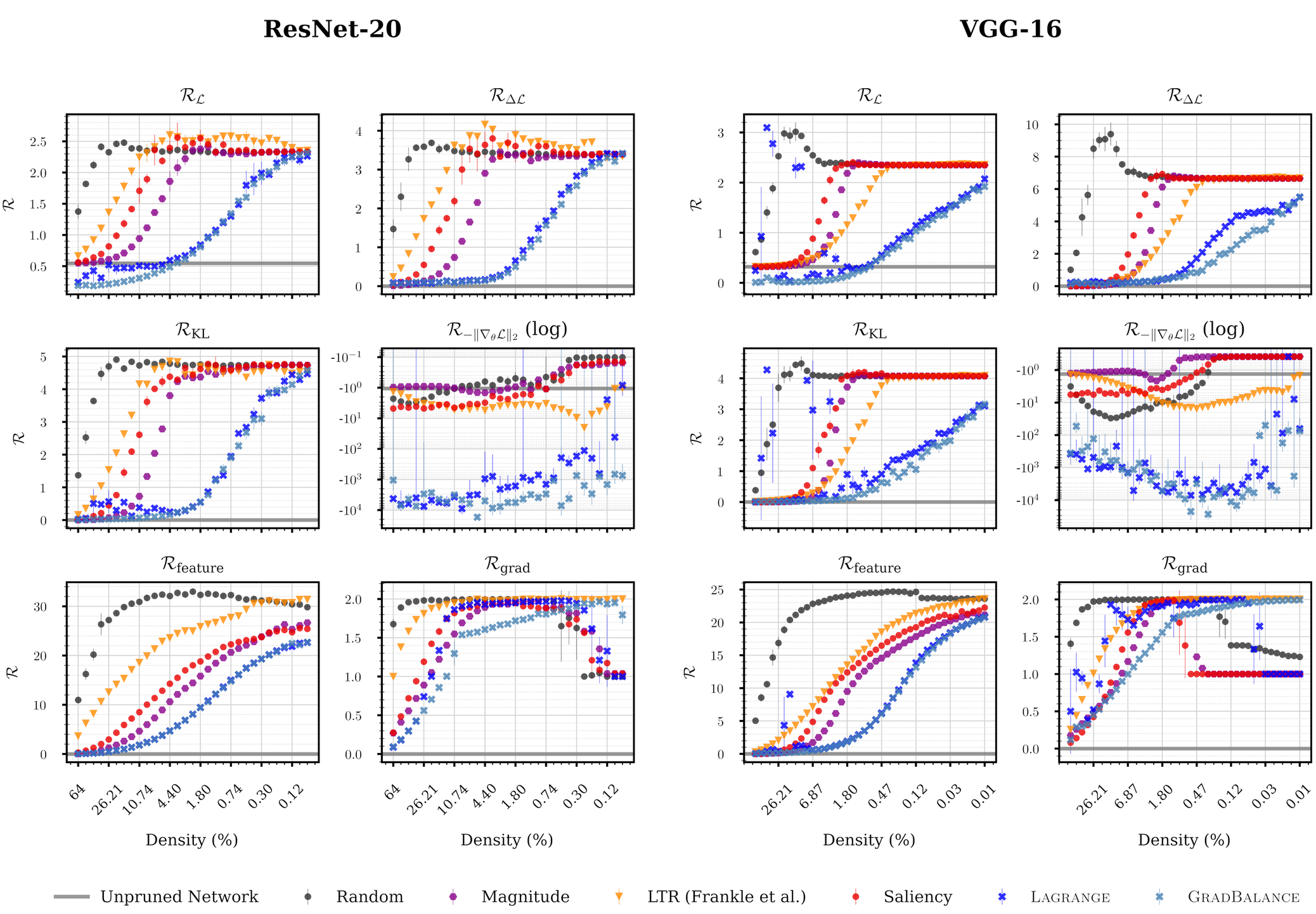}
    \caption{Objective performance with respect to sparsity of subnetworks of ResNet-20 (top) and VGG-16 (bottom) on CIFAR10 drawn using respective objective functions (as outlined in Section \ref{ss:choosingtheobjectivefunction}). Presented results are computed immediately after the ticket is drawn; here, we apply all methods at the rewinding iteration. We plot baseline results from random pruning, magnitude pruning, LTR, and saliency pruning. We include results under both the \textsc{Lagrange} and \textsc{GradBalance} gradient steps, and all results are under Quick CTS. Both versions of Quick CTS overwhelmingly excel on all non-defective objectives, with \textsc{GradBalance}-based tickets performing greater and with more consistency than \textsc{Lagrange}-based ones.}\label{fig:objectives}
\end{figure*}

\subsection{Objective Performance on CIFAR-10}\label{ss:objectiveperformance}
In Section~\ref{s:issuewithsaliency}, we illustrated the poor performance of saliency-based pruning methods on objective performance at initialization. Here, we expand on this by presenting the results over all 6 objective functions of the proposed method. These are computed at the rewinding iteration, i.e., immediately after ticket drawing. For completeness, we also present results at initialization for \(\mathcal{R}_{\mathcal{L}}\) and \(\mathcal{R}_{-\lVert \nabla_{\theta} \mathcal{L} \rVert_2}\) at initialization, which we denote with overhead bars (i.e., \(\overline{\mathcal{R}_{\mathcal{L}}}\)). All of our presented results are utilizing Quick CTS, which is computed in only $1/8$ of training time. We consider the results of both gradient step algorithms, \textsc{Lagrange} (Algorithm~\ref{lagrangemultiplieralgoirthm}) and \textsc{GradBalance} (Algorithm~\ref{gradbalancealgorithm}). As baselines, we also consider the results of LTR~\cite{instability}, saliency-based pruning (i.e., SNIP~\cite{snip} for \(\mathcal{R}_{\mathcal{L}}\) and \(\mathcal{R}_{\Delta \mathcal{L}}\), GraSP~\cite{grasp} for \(\mathcal{R}_{-\lVert \nabla_{\theta} \mathcal{L} \rVert_2}\), and similarly designed pruning procedures for the rest), random pruning, and magnitude-based pruning. 

However, saliency-based approaches are not designed for our knowledge-distillation-inspired objectives. As explained in Section \ref{s:issuewithsaliency}, they assign importance scores by calculating an objective function \textit{over the dense network}, and then considering how large a contribution each weight has on the gradient of the network. However, when the student is equal to the dense network, all such deviance-measuring objectives are perfectly optimized, and the gradient over the entire network is zero. To account for this, we instead implement the following procedure. Define an overlay \(s := \boldsymbol{1}_d\), apply a small amount of Gaussian noise to it (we use \(\sigma = 6 \cdot 10^{-2}\)), and compute the objective value with \(\theta_{\text{eff}} = s \odot \theta \). We then compute \(S(\theta) = \frac{\partial \mathcal{R}}{\partial s}\), since \(\frac{\partial \mathcal{R}}{\partial s_j} = \theta_j \frac{\partial \mathcal{R}}{\partial \theta_j}\). This thus allows for meaningful, nonzero gradient calculation. Additionally, motivated by \cite{pruningatinitialization}, we opt to prune the scores of least magnitude, rather than pruning by raw score (i.e., let \(S(\theta)_{\text{eff}} = \left|S(\theta)\right|\)). 

Fig.~\ref{fig:objectivesinit} shows the results for select methods that prune at initialization. Although ticket search dynamics in this period can be relatively unstable, the proposed method performs significantly better than all other methods over all sparsities. We plot \(\overline{\mathcal{R}_{-\lVert \nabla_{\theta} \mathcal{L} \rVert_{2}}}\) on a log scale for visual clarity; our method draws tickets with orders of magnitude larger gradient norms than others. Additionally, particularly for \(\overline{\mathcal{R}_{\mathcal{L}}}\) on ResNet-20, the need for gradient balancing is underscored. Even with extensive tuning over the relative learning strength of the dual optimization with the \textsc{Lagrange} gradient step, it suffers extreme instability due to the rigid nature of the search space at initialization.

Fig.~\ref{fig:objectives} shows the results for both ResNet-20 and VGG-16. As with the tickets drawn at initialization, the proposed method performs significantly better than all other methods over all sparsities. The one caveat to this occurs for \(\mathcal{R}_{\text{grad}}\), which is seemingly deficient in its ability to assess subnetworks, likely due to excessive noise. Here, the instability of \textsc{Lagrange} remains visible, although it is still able to mostly outperform all baselines. 

Interestingly, aside from \(\mathcal{R}_{-\lVert \nabla_{\theta} \mathcal{L} \rVert_{2}}\), on ResNet-20, LTR performs worse on all objectives than all other methods outside of random pruning. This occurs even in sparsity regimes where it boasts SOTA accuracy, and is not apparent on VGG-16 (outside of the noisy \(\mathcal{R}_{\text{feature}}\)). This seems to indicate that on ResNet-20 on CIFAR-10, LTR tickets do not preserve the training trajectory of their parent network, even if they converge to a similar loss basin.

\renewcommand{\arraystretch}{1.25}
\begin{table}
\caption{Comparison of CTS vs LTR subnetwork performance and total runtime to achieve said performance across select sparsities. Experimental configurations can be found in Section \ref{ss:experimentalconfig}. We report higher-k accuracy values for ResNet-50 at more extreme sparsity levels to maintain readability}
\label{tab:performance-time-resnet50}
\centering
\resizebox{\columnwidth}{!}{%
\renewcommand{\arraystretch}{1.25}
\begin{tabular}{c c | c | cc | cc}
\hline
\textbf{Model} & \textbf{Sparsity} & \textbf{Metric} &
\multicolumn{2}{c|}{\textbf{CTS\textsubscript{KL}}} &
\multicolumn{2}{c}{\textbf{LTR}} \\
& & & \textbf{Perf.} & \textbf{Time} & \textbf{Perf.} & \textbf{Time} \\
\hline
\multirow{3}{*}{ResNet-20} & 98.2\% & \multirow{3}{*}{acc@1} & 81.7\% & 7.9 min & 83.5\% & 78.7 min \\
                            & 99.3\% &                          & 74.0\% & 7.9 min & 68.3\% & 95.2 min \\
                            & 99.7\% &                          & 58.8\% & 7.9 min & 55.7\% & 108.3 min \\
\hline
\multirow{3}{*}{ResNet-50}
  & 90.0\% & acc@1 & \(66.3\%\) & \(20.7\) hr & \(71.3\%\) & \(35.2\) hr \\
  & 99.0\% & acc@3 & \(65.4\%\) & \(20.8\) hr & \(60.5\%\) & \(59.1\) hr \\
  & 99.9\% & acc@5 & \(24.2\%\) & \(20.7\) hr & \(6.7\%\) & \(82.6\) hr \\
\hline
\end{tabular}%
}
\end{table}

\subsection{Performance-Time Comparison}
We conclude by presenting a subnetwork performance vs. time comparison of CTS\textsubscript{KL} against LTR, the current state-of-the-art. Times presented include both ticket drawing and training times. Table \ref{tab:performance-time-resnet50} displays this comparison for select sparsities for two tasks: ResNet-50 on ImageNet and ResNet-20 on CIFAR-10. Our baseline accuracy for ResNet-50 on ImageNet is \(74.77\%\), and for ResNet-20 on CIFAR-10 is \(92.12\%\). The results presented follow the experimental configuration outlined in Section \ref{ss:experimentalconfig}. Notably, on ResNet-50, LTR has a significantly increased pruning ratio (\(53.58\%\) per iteration rather than \(20\%\)) and ticket search is carried out with half of the total training epochs, i.e., 45 rather than 90. In our implementation, we do not utilize the speedups of sparse training, largely due to the lack of support among current libraries. 

Though more apparent on ResNet-20 because of the aforementioned changes, CTS computes winning subnetworks significantly faster than LTR on both tasks. It consistently achieves similar or better task performance, though its benefits are clearly more visible in the sparse regime. It also avoids the iterative nature of LTR; tickets of all sparsities can be drawn in equal time.

\section{Conclusion}
In this article, we identified a fundamental limitation in existing Pruning-at-Initialization methods: their dependence on first-order saliency scores fails to capture the complex, interdependent nature of neural network weights, leading to suboptimal performance, especially at high sparsity. To address this, we proposed CTS, a new paradigm for finding lottery tickets that abandons local saliency metrics in favor of a holistic, probabilistic search over the entire subnetwork.

Our method frames ticket discovery as a constrained optimization problem, made tractable through a concrete relaxation and a novel, hyperparameter-free gradient balancing algorithm for precise sparsity control. By performing this search early in training rather than strictly at initialization, and by utilizing objective functions inspired by knowledge distillation, we can effectively preserve the training dynamics of the original dense network. Our experimental results demonstrate that CTS, particularly when guided by a reverse KL divergence objective (\(\textsc{CTS}_{\text{KL}}\)), sets a new state-of-the-art in the high-sparsity regime. It outperforms the computationally expensive LTR baseline in the highly sparse regime and remains competitive among all models. For example, CTS\textsubscript{KL} finds subnetworks 3 times faster than LTR (with a reduced pruning rate) at \(99.0\%\) sparsity on ResNet-50 on ImageNet and 12 times faster than LTR at \(99.3\%\) sparsity on ResNet-20 on CIFAR-10. At the aforementioned sparsities, CTS\textsubscript{KL} achieves a top-3 accuracy of \(65.4\%\) in comparison to LTR's \(60.5\%\) on ResNet-50 and a top-1 accuracy of \(74.0\%\) in comparison to LTR's \(68.3\%\) on ResNet-20. Furthermore, the tickets generated through CTS successfully pass the sanity checks that reveal the shortcomings of saliency-based PaI methods.

This work represents a significant step in providing a general framework for drawing tickets that avoid the pitfalls of such sanity checks. However, the tickets drawn through CTS under both the \textsc{GradBalance} and \textsc{Lagrange} gradient steps remain unable to match LTR performance at the matching sparsities, i.e., sparsity ranges in which LTR is able to produce subnetworks with \textit{no} performance decline. While not nearly as sparse as those drawn in this article, these subnetworks have a similarly high demand in many use cases. Future work will be dedicated to the search for algorithms that can draw matching subnetworks more efficiently than LTR. 

Additionally, nearly all work on the lottery ticket hypothesis currently remains largely confined to image classification tasks. It remains to be seen whether the advantages of CTS can translate to other tasks like NLP, particularly with the additional complexities of self-attention. 

\section*{Acknowledgment}

This material is based on work supported by the National Science Foundation under grant no. 2318139 and by the Daimler Truck Portland professorship.

\bibliographystyle{IEEEtran}
\bibliography{references}

@inproceedings{lotterytickethypothesis,
title={The Lottery Ticket Hypothesis: Finding Sparse, Trainable Neural Networks},
author={Jonathan Frankle and Michael Carbin},
booktitle={International Conference on Learning Representations},
year={2019},
url={https://openreview.net/forum?id=rJl-b3RcF7},
}

@inproceedings{instability,
  title = 	 {Linear Mode Connectivity and the Lottery Ticket Hypothesis},
  author =       {Frankle, Jonathan and Dziugaite, Gintare Karolina and Roy, Daniel and Carbin, Michael},
  booktitle = 	 {Proceedings of the 37th International Conference on Machine Learning},
  pages = 	 {3259--3269},
  year = 	 {2020},
  editor = 	 {III, Hal Daumé and Singh, Aarti},
  volume = 	 {119},
  series = 	 {Proceedings of Machine Learning Research},
  month = 	 {13--18 Jul},
  publisher =    {PMLR},
  url = 	 {https://proceedings.mlr.press/v119/frankle20a.html}
}

@article{evci_gradient_flow, title={Gradient Flow in Sparse Neural Networks and How Lottery Tickets Win}, volume={36}, url={https://ojs.aaai.org/index.php/AAAI/article/view/20611}, DOI={10.1609/aaai.v36i6.20611}, abstractNote={Sparse Neural Networks (NNs) can match the generalization of dense NNs using a fraction of the compute/storage for inference, and have the potential to enable efficient training. However, naively training unstructured sparse NNs from random initialization results in significantly worse generalization, with the notable exceptions of Lottery Tickets (LTs) and Dynamic Sparse Training (DST). In this work, we attempt to answer: (1) why training unstructured sparse networks from random initialization performs poorly and; (2) what makes LTs and DST the exceptions? We show that sparse NNs have poor gradient flow at initialization and propose a modified initialization for unstructured connectivity. Furthermore, we find that DST methods significantly improve gradient flow during training over traditional sparse training methods. Finally, we show that LTs do not improve gradient flow, rather their success lies in re-learning the pruning solution they are derived from — however, this comes at the cost of learning novel solutions.}, number={6}, journal={Proceedings of the AAAI Conference on Artificial Intelligence}, author={Evci, Utku and Ioannou, Yani and Keskin, Cem and Dauphin, Yann}, year={2022}, month={Jun.}, pages={6577-6586} }

@inproceedings{ pruningatinitialization,
 author       = {Jonathan Frankle and
                  Gintare Karolina Dziugaite and
                  Daniel M. Roy and
                  Michael Carbin},
  title        = {Pruning Neural Networks at Initialization: Why Are We Missing the
                  Mark?},
  booktitle    = {9th International Conference on Learning Representations, {ICLR} 2021,
                  Virtual Event, Austria, May 3-7, 2021},
  publisher    = {OpenReview.net},
  year         = {2021},
  url          = {https://openreview.net/forum?id=Ig-VyQc-MLK},
  bibsource    = {dblp computer science bibliography, https://dblp.org}
}

@inproceedings{
snip,
title={{SNIP}: {SINGLE}-{SHOT} {NETWORK} {PRUNING} {BASED} {ON} {CONNECTION} {SENSITIVITY}},
author={Namhoon Lee and Thalaiyasingam Ajanthan and Philip Torr},
booktitle={International Conference on Learning Representations},
year={2019},
url={https://openreview.net/forum?id=B1VZqjAcYX},
}

@inproceedings{
grasp,
title={Picking Winning Tickets Before Training by Preserving Gradient Flow},
author={Chaoqi Wang and Guodong Zhang and Roger Grosse},
booktitle={International Conference on Learning Representations},
year={2020},
url={https://openreview.net/forum?id=SkgsACVKPH}
}

@inproceedings{
 neuraltangentkernel,
 author = {Jacot, Arthur and Gabriel, Franck and Hongler, Clement},
 booktitle = {Advances in Neural Information Processing Systems},
 editor = {S. Bengio and H. Wallach and H. Larochelle and K. Grauman and N. Cesa-Bianchi and R. Garnett},
 pages = {8917--8927},
 publisher = {Curran Associates, Inc.},
 title = {Neural Tangent Kernel: Convergence and Generalization in Neural Networks},
 url = {https://proceedings.neurips.cc/paper_files/paper/2018/file/5a4be1fa34e62bb8a6ec6b91d2462f5a-Paper.pdf},
 volume = {31},
 year = {2018}
}

@inproceedings{
synflow,
 author = {Tanaka, Hidenori and Kunin, Daniel and Yamins, Daniel L and Ganguli, Surya},
 booktitle = {Advances in Neural Information Processing Systems},
 editor = {H. Larochelle and M. Ranzato and R. Hadsell and M.F. Balcan and H. Lin},
 pages = {6377--6389},
 publisher = {Curran Associates, Inc.},
 title = {Pruning neural networks without any data by iteratively conserving synaptic flow},
 url = {https://proceedings.neurips.cc/paper_files/paper/2020/file/46a4378f835dc8040c8057beb6a2da52-Paper.pdf},
 volume = {33},
 year = {2020}
}

@inproceedings{
learningl0sparse,
title={Learning Sparse Neural Networks through $L_0$ Regularization},
author={Christos Louizos and Max Welling and Diederik P. Kingma},
booktitle={International Conference on Learning Representations},
year={2018},
url={https://openreview.net/forum?id=H1Y8hhg0b},
}

@inproceedings{
learningraterewinding,
title={Comparing Rewinding and Fine-tuning in Neural Network Pruning},
author={Alex Renda and Jonathan Frankle and Michael Carbin},
booktitle={International Conference on Learning Representations},
year={2020},
url={https://openreview.net/forum?id=S1gSj0NKvB}
}

@inproceedings{
whylearningraterewinding,
title={Masks, Signs, And Learning Rate Rewinding},
author={Advait Harshal Gadhikar and Rebekka Burkholz},
booktitle={The Twelfth International Conference on Learning Representations},
year={2024},
url={https://openreview.net/forum?id=qODvxQ8TXW}
}

@inproceedings{
earlyphase,
title={The Early Phase of Neural Network Training},
author={Jonathan Frankle and David J. Schwab and Ari S. Morcos},
booktitle={International Conference on Learning Representations},
year={2020},
url={https://openreview.net/forum?id=Hkl1iRNFwS}
}

@inproceedings{
butterflyeffect,
title={The Butterfly Effect: Neural Network Training Trajectories Are Highly Sensitive to Initial Conditions},
author={G{\"u}l Sena Alt{\i}nta{\c{s}} and Devin Kwok and Colin Raffel and David Rolnick},
booktitle={Forty-second International Conference on Machine Learning},
year={2025},
url={https://openreview.net/forum?id=L1Bm396P0X}
}

@inproceedings{
  hessianeigenvalue,
  title = 	 {An Investigation into Neural Net Optimization via Hessian Eigenvalue Density},
  author =       {Ghorbani, Behrooz and Krishnan, Shankar and Xiao, Ying},
  booktitle = 	 {Proceedings of the 36th International Conference on Machine Learning},
  pages = 	 {2232--2241},
  year = 	 {2019},
  editor = 	 {Chaudhuri, Kamalika and Salakhutdinov, Ruslan},
  volume = 	 {97},
  series = 	 {Proceedings of Machine Learning Research},
  month = 	 {09--15 Jun},
  publisher =    {PMLR},
  url = 	 {https://proceedings.mlr.press/v97/ghorbani19b.html}
}

@inproceedings{
criticallearningperiods,
title={Critical Learning Periods Emerge Even in Deep Linear Networks},
author={Michael Kleinman and Alessandro Achille and Stefano Soatto},
booktitle={The Twelfth International Conference on Learning Representations},
year={2024},
url={https://openreview.net/forum?id=Aq35gl2c1k}
}

@inproceedings{
edgepopup,
author = {Ramanujan, Vivek and Wortsman, Mitchell and Kembhavi, Aniruddha and Farhadi, Ali and Rastegari, Mohammad},
pages={11890--11899},
title = {What's Hidden in a Randomly Weighted Neural Network?},
booktitle = {Proceedings of the IEEE/CVF Conference on Computer Vision and Pattern Recognition (CVPR)},
month = {June},
year = {2020}
}

@inproceedings{
 gemminer,
 author = {Sreenivasan, Kartik and Sohn, Jy-yong and Yang, Liu and Grinde, Matthew and Nagle, Alliot and Wang, Hongyi and Xing, Eric and Lee, Kangwook and Papailiopoulos, Dimitris},
 booktitle = {Advances in Neural Information Processing Systems},
 editor = {S. Koyejo and S. Mohamed and A. Agarwal and D. Belgrave and K. Cho and A. Oh},
 pages = {14529--14540},
 publisher = {Curran Associates, Inc.},
 title = {Rare Gems: Finding Lottery Tickets at Initialization},
 url = {https://proceedings.neurips.cc/paper_files/paper/2022/file/5d52b102ebd672023628cac20e9da5ff-Paper-Conference.pdf},
 volume = {35},
 year = {2022}
}

@inproceedings{
  overparameterizationacceleration,
  title = 	 {On the Optimization of Deep Networks: Implicit Acceleration by Overparameterization},
  author =       {Arora, Sanjeev and Cohen, Nadav and Hazan, Elad},
  booktitle = 	 {Proceedings of the 35th International Conference on Machine Learning},
  pages = 	 {244--253},
  year = 	 {2018},
  editor = 	 {Dy, Jennifer and Krause, Andreas},
  volume = 	 {80},
  series = 	 {Proceedings of Machine Learning Research},
  month = 	 {10--15 Jul},
  publisher =    {PMLR},
  url = 	 {https://proceedings.mlr.press/v80/arora18a.html}
}

@article{overparameterizationgeneralization,
author = {Zhang, Chiyuan and Bengio, Samy and Hardt, Moritz and Recht, Benjamin and Vinyals, Oriol},
title = {Understanding deep learning (still) requires rethinking generalization},
year = {2021},
issue_date = {March 2021},
publisher = {Association for Computing Machinery},
address = {New York, NY, USA},
volume = {64},
number = {3},
issn = {0001-0782},
url = {https://doi.org/10.1145/3446776},
doi = {10.1145/3446776},
journal = {Commun. ACM},
month = feb,
pages = {107–115},
numpages = {9}
}

@inproceedings{
minillm,
title={Mini{LLM}: Knowledge Distillation of Large Language Models},
author={Yuxian Gu and Li Dong and Furu Wei and Minlie Huang},
booktitle={The Twelfth International Conference on Learning Representations},
year={2024},
url={https://openreview.net/forum?id=5h0qf7IBZZ}
}

@inproceedings{rethinkingkld,
  title     = {Rethinking {K}ullback--{L}eibler Divergence in Knowledge Distillation for Large Language Models},
  author    = {Wu, Taiqiang and Tao, Chaofan and Wang, Jiahao and Yang, Runming and Zhao, Zhe and Wong, Ngai},
  booktitle = {Proceedings of the 31st International Conference on Computational Linguistics},
  address   = {Abu Dhabi, UAE},
  month     = jan,
  year      = {2025},
  pages     = {5737--5755},
  publisher = {Association for Computational Linguistics},
  url       = {https://aclanthology.org/2025.coling-main.383/}
}

@article{selfdistillationdropout,
author = {Lee, Hyoje and Park, Yeachan and Seo, Hyun and Kang, Myungjoo},
title = {Self-knowledge distillation via dropout},
year = {2023},
issue_date = {Aug 2023},
publisher = {Elsevier Science Inc.},
address = {USA},
volume = {233},
number = {C},
issn = {1077-3142},
url = {https://doi.org/10.1016/j.cviu.2023.103720},
doi = {10.1016/j.cviu.2023.103720},
journal = {Comput. Vis. Image Underst.},
month = aug,
numpages = {9}
}

@inproceedings{
  angulardistillation,
  author={Liu, Tao and Chen, Chenshu and Yang, Xi and Tan, Wenming},
  booktitle={2024 IEEE/CVF Winter Conference on Applications of Computer Vision (WACV)}, 
  title={Rethinking Knowledge Distillation with Raw Features for Semantic Segmentation}, 
  year={2024},
  volume={},
  number={},
  pages={1144-1153},
  doi={10.1109/WACV57701.2024.00119}
}

@inproceedings{
fitnets,
title={FitNets: Hints for Thin Deep Nets},
author={Adriana Romero and Nicolas Ballas and Samira Ebrahimi Kahou and Antoine Chassang and Carlo Gatta and Yoshua Bengio},
booktitle={International Conference on Learning Representations},
year={2015},
url={https://arxiv.org/abs/1412.6550}
}

@InProceedings{gradnorm,
  title = 	 {{G}rad{N}orm: Gradient Normalization for Adaptive Loss Balancing in Deep Multitask Networks},
  author =       {Chen, Zhao and Badrinarayanan, Vijay and Lee, Chen-Yu and Rabinovich, Andrew},
  booktitle = 	 {Proceedings of the 35th International Conference on Machine Learning},
  pages = 	 {794--803},
  year = 	 {2018},
  editor = 	 {Dy, Jennifer and Krause, Andreas},
  volume = 	 {80},
  series = 	 {Proceedings of Machine Learning Research},
  month = 	 {10--15 Jul},
  publisher =    {PMLR},
  url = 	 {https://proceedings.mlr.press/v80/chen18a.html}
}

@inproceedings{
concretedistribution,
title={The Concrete Distribution: A Continuous Relaxation of Discrete Random Variables},
author={Chris J. Maddison and Andriy Mnih and Yee Whye Teh},
booktitle={International Conference on Learning Representations},
year={2017},
url={https://openreview.net/forum?id=S1jE5L5gl}
}

@article{
  estimatorreinforce,
  author    = {Williams, Ronald J.},
  title     = {Simple statistical gradient-following algorithms for connectionist reinforcement learning},
  journal   = {Machine Learning},
  year      = {1992},
  volume    = {8},
  number    = {3},
  pages     = {229--256},
  doi       = {10.1007/BF00992696},
  url       = {https://doi.org/10.1007/BF00992696},
  issn      = {1573-0565}
}

@inproceedings{estimatorrebar,
 author = {Tucker, George and Mnih, Andriy and Maddison, Chris J and Lawson, John and Sohl-Dickstein, Jascha},
 booktitle = {Advances in Neural Information Processing Systems},
 editor = {I. Guyon and U. Von Luxburg and S. Bengio and H. Wallach and R. Fergus and S. Vishwanathan and R. Garnett},
 pages = {2627--2636},
 publisher = {Curran Associates, Inc.},
 title = {REBAR: Low-variance, unbiased gradient estimates for discrete latent variable models},
 url = {https://proceedings.neurips.cc/paper_files/paper/2017/file/ebd6d2f5d60ff9afaeda1a81fc53e2d0-Paper.pdf},
 volume = {30},
 year = {2017}
}

@inproceedings{
estimatorrelax,
title={Backpropagation through the Void: Optimizing control variates for black-box gradient estimation},
author={Will Grathwohl and Dami Choi and Yuhuai Wu and Geoff Roeder and David Duvenaud},
booktitle={International Conference on Learning Representations},
year={2018},
url={https://openreview.net/forum?id=SyzKd1bCW},
}

@inproceedings{
estimatorarm,
title={{ARM}: Augment-{REINFORCE}-Merge Gradient for Stochastic Binary Networks},
author={Mingzhang Yin and Mingyuan Zhou},
booktitle={International Conference on Learning Representations},
year={2019},
url={https://openreview.net/forum?id=S1lg0jAcYm},
}

@article{estimatorste,
       author = {{Bengio}, Yoshua},
        title = "{Estimating or Propagating Gradients Through Stochastic Neurons}",
      journal = {arXiv e-prints},
         year = 2013,
        month = may,
          eid = {arXiv:1305.2982},
        pages = {arXiv:1305.2982},
          doi = {10.48550/arXiv.1305.2982},
archivePrefix = {arXiv},
       eprint = {1305.2982},
 primaryClass = {cs.LG},
       adsurl = {https://ui.adsabs.harvard.edu/abs/2013arXiv1305.2982B}
}

@inproceedings{
proxsgd,
title={ProxSGD: Training Structured Neural Networks under Regularization and Constraints},
author={Yang Yang and Yaxiong Yuan and Avraam Chatzimichailidis and Ruud JG van Sloun and Lei Lei and Symeon Chatzinotas},
booktitle={International Conference on Learning Representations},
year={2020},
url={https://openreview.net/forum?id=HygpthEtvr}
}

@inproceedings{sanitychecks,
 author = {Su, Jingtong and Chen, Yihang and Cai, Tianle and Wu, Tianhao and Gao, Ruiqi and Wang, Liwei and Lee, Jason D},
 booktitle = {Advances in Neural Information Processing Systems},
 editor = {H. Larochelle and M. Ranzato and R. Hadsell and M.F. Balcan and H. Lin},
 pages = {20390--20401},
 publisher = {Curran Associates, Inc.},
 title = {Sanity-Checking Pruning Methods: Random Tickets can Win the Jackpot},
 url = {https://proceedings.neurips.cc/paper_files/paper/2020/file/eae27d77ca20db309e056e3d2dcd7d69-Paper.pdf},
 volume = {33},
 year = {2020}
}

@inproceedings{han2015compression,
  author       = {Song Han and
                  Huizi Mao and
                  William J. Dally},
  editor       = {Yoshua Bengio and
                  Yann LeCun},
  title        = {Deep Compression: Compressing Deep Neural Network with Pruning, Trained
                  Quantization and Huffman Coding},
  booktitle    = {4th International Conference on Learning Representations, {ICLR} 2016,
                  San Juan, Puerto Rico, May 2-4, 2016, Conference Track Proceedings},
  year         = {2016},
  url          = {http://arxiv.org/abs/1510.00149},
  bibsource    = {dblp computer science bibliography, https://dblp.org}
}

@inproceedings{optimalbrainsurgeon,
 author = {Hassibi, Babak and Stork, David},
 booktitle = {Advances in Neural Information Processing Systems},
 editor = {S. Hanson and J. Cowan and C. Giles},
 pages = {164--171},
 publisher = {Morgan-Kaufmann},
 title = {Second order derivatives for network pruning: Optimal Brain Surgeon},
 url = {https://proceedings.neurips.cc/paper_files/paper/1992/file/303ed4c69846ab36c2904d3ba8573050-Paper.pdf},
 volume = {5},
 year = {1992}
}

@inproceedings{ supermasks,
 author = {Zhou, Hattie and Lan, Janice and Liu, Rosanne and Yosinski, Jason},
 booktitle = {Advances in Neural Information Processing Systems},
 editor = {H. Wallach and H. Larochelle and A. Beygelzimer and F. d\textquotesingle Alch\'{e}-Buc and E. Fox and R. Garnett},
 pages = {3597--3607},
 publisher = {Curran Associates, Inc.},
 title = {Deconstructing Lottery Tickets: Zeros, Signs, and the Supermask},
 url = {https://proceedings.neurips.cc/paper_files/paper/2019/file/1113d7a76ffceca1bb350bfe145467c6-Paper.pdf},
 volume = {32},
 year = {2019}
}

@inproceedings{
snowsprune,
title={Preserving Deep Representations in One-Shot Pruning: A Hessian-Free Second-Order Optimization Framework},
author={Ryan Lucas and Rahul Mazumder},
booktitle={The Thirteenth International Conference on Learning Representations},
year={2025},
url={https://openreview.net/forum?id=eNQp79A5Oz}
}

@inproceedings{chitaprune,
  title = 	 {Fast as {CHITA}: Neural Network Pruning with Combinatorial Optimization},
  author =       {Benbaki, Riade and Chen, Wenyu and Meng, Xiang and Hazimeh, Hussein and Ponomareva, Natalia and Zhao, Zhe and Mazumder, Rahul},
  booktitle = 	 {Proceedings of the 40th International Conference on Machine Learning},
  pages = 	 {2031--2049},
  year = 	 {2023},
  editor = 	 {Krause, Andreas and Brunskill, Emma and Cho, Kyunghyun and Engelhardt, Barbara and Sabato, Sivan and Scarlett, Jonathan},
  volume = 	 {202},
  series = 	 {Proceedings of Machine Learning Research},
  month = 	 {23--29 Jul},
  publisher =    {PMLR},
  url = 	 {https://proceedings.mlr.press/v202/benbaki23a.html}
}

@InProceedings{spodeprune,
  title = 	 {Pruning via Sparsity-indexed {ODE}: a Continuous Sparsity Viewpoint},
  author =       {Mo, Zhanfeng and Shi, Haosen and Pan, Sinno Jialin},
  booktitle = 	 {Proceedings of the 40th International Conference on Machine Learning},
  pages = 	 {25018--25036},
  year = 	 {2023},
  editor = 	 {Krause, Andreas and Brunskill, Emma and Cho, Kyunghyun and Engelhardt, Barbara and Sabato, Sivan and Scarlett, Jonathan},
  volume = 	 {202},
  series = 	 {Proceedings of Machine Learning Research},
  month = 	 {23--29 Jul},
  publisher =    {PMLR},
  url = 	 {https://proceedings.mlr.press/v202/mo23c.html}
}

@Techreport{cifar10,
 author = {Krizhevsky, Alex and Hinton, Geoffrey},
 address = {Toronto, Ontario},
 institution = {University of Toronto},
 number = {0},
 publisher = {Technical report, University of Toronto},
 title = {Learning multiple layers of features from tiny images},
 year = {2009},
 title_with_no_special_chars = {Learning multiple layers of features from tiny images},
 url = {https://www.cs.toronto.edu/~kriz/learning-features-2009-TR.pdf}
}

@inproceedings{resnet20,
  author={He, Kaiming and Zhang, Xiangyu and Ren, Shaoqing and Sun, Jian},
  booktitle={2016 IEEE Conference on Computer Vision and Pattern Recognition (CVPR)}, 
  title={Deep Residual Learning for Image Recognition}, 
  year={2016},
  volume={},
  number={},
  pages={770-778},
  doi={10.1109/CVPR.2016.90}}

@inproceedings{vgg16,
  author       = "Karen Simonyan and Andrew Zisserman",
  title        = "Very Deep Convolutional Networks for Large-Scale Image Recognition",
  pages = {1--14}, 
  publisher = "Computational and Biological Learning Society",
  booktitle    = "3rd International Conference on Learning Representations",
  year         = "2015",
}

@article{imagenet,
Author = {Olga Russakovsky and Jia Deng and Hao Su and Jonathan Krause and Sanjeev Satheesh and Sean Ma and Zhiheng Huang and Andrej Karpathy and Aditya Khosla and Michael Bernstein and Alexander C. Berg and Li Fei-Fei},
Title = {{ImageNet Large Scale Visual Recognition Challenge}},
Year = {2015},
journal   = {International Journal of Computer Vision (IJCV)},
doi = {10.1007/s11263-015-0816-y},
volume={115},
number={3},
pages={211-252}
}

@misc{openlth,
  author       = "Frankle, Jonathan",
  title        = "{OpenLTH: A Framework for Lottery Tickets and Beyond}",
  howpublished = "\url{https://github.com/facebookresearch/open_lth}",
}

@article{equivariantcircuits,
  author    = {Olah, Chris and Cammarata, Nick and Voss, Chelsea and Schubert, Ludwig and Goh, Gabriel},
  title     = {Naturally Occurring Equivariance in Neural Networks},
  journal   = {Distill},
  year      = {2020},
  month     = dec,
  day       = {8},
  note      = {Part of the “Circuits” thread},
  doi       = {10.23915/distill.00024.004},
  url       = {https://distill.pub/2020/circuits/equivariance/}
}

@inproceedings{
correlationmodedecomp,
title={Enhancing Neural Training via a Correlated Dynamics Model},
author={Jonathan Brokman and Roy Betser and Rotem Turjeman and Tom Berkov and Ido Cohen and Guy Gilboa},
booktitle={The Twelfth International Conference on Learning Representations},
year={2024},
url={https://openreview.net/forum?id=c9xsaASm9L}
}

@article{edgedevices,
author = {Hoffpauir, Kyle and Simmons, Jacob and Schmidt, Nikolas and Pittala, Rachitha and Briggs, Isaac and Makani, Shanmukha and Jararweh, Yaser},
title = {A Survey on Edge Intelligence and Lightweight Machine Learning Support for Future Applications and Services},
year = {2023},
issue_date = {June 2023},
publisher = {Association for Computing Machinery},
address = {New York, NY, USA},
volume = {15},
number = {2},
issn = {1936-1955},
url = {https://doi.org/10.1145/3581759},
doi = {10.1145/3581759},
journal = {J. Data and Information Quality},
month = jun,
articleno = {20},
numpages = {30}
}

@inproceedings{patheXclusion,
  author={Iurada, Leonardo and Ciccone, Marco and Tommasi, Tatiana},
  booktitle={2024 IEEE/CVF Conference on Computer Vision and Pattern Recognition (CVPR)}, 
  title={Finding Lottery Tickets in Vision Models via Data-Driven Spectral Foresight Pruning}, 
  year={2024},
  volume={},
  number={},
  pages={16142-16151},
  doi={10.1109/CVPR52733.2024.01528}}

@inproceedings{epsd,
author = {Chen, Dong and Liu, Ning and Zhu, Yichen and Che, Zhengping and Ma, Rui and Zhang, Fachao and Mou, Xiaofeng and Chang, Yi and Tang, Jian},
title = {EPSD: early pruning with self-distillation for efficient model compression},
year = {2024},
isbn = {978-1-57735-887-9},
publisher = {AAAI Press},
url = {https://doi.org/10.1609/aaai.v38i10.29004},
doi = {10.1609/aaai.v38i10.29004},
booktitle = {Proceedings of the Thirty-Eighth AAAI Conference on Artificial Intelligence and Thirty-Sixth Conference on Innovative Applications of Artificial Intelligence and Fourteenth Symposium on Educational Advances in Artificial Intelligence},
articleno = {1256},
numpages = {9},
series = {AAAI'24}
}

\end{document}